\documentclass{article}

\usepackage{jfrExamplee}
\usepackage[letterpaper,margin=1in]{geometry}
\usepackage{amsmath,amsthm,amsfonts,amscd} 
\usepackage{amssymb}
\usepackage{citesort}
\usepackage{url}
\usepackage{bm}
\usepackage{graphicx}
\usepackage{cancel}
\usepackage{subfigure} 
\usepackage{setspace}
\usepackage{caption}
\usepackage{natbib}
\newcommand{\ve}[1]{\ensuremath{{\bf #1}}}
\newcommand{\vet}[1]{\ensuremath{{\bf #1}(t)}}

\newcommand{\norm}[1]{\ensuremath{{\left|\left|{#1}\right|\right|}}}
\newcommand{\tny}[1] {\ensuremath{{\scriptscriptstyle #1}}}

\renewcommand{\norm}[1]{\left| \left| #1 \right| \right|}

\newcommand{\sth}{\sin{\theta}}
\newcommand{\cth}{\cos{\theta}}
\newcommand{\Eq}[1]{Equation~(\ref{#1})}
\newcommand{\Sec}[1]{Section~\ref{#1}}
\newcommand{\Fig}[1]{Figure~\ref{#1}}

\author{
Shilpa Gulati\thanks{{\bf Corresponding author}. This work was performed as part of Shilpa's Ph.D.~\citep{Gulati_2011} in the Mechanical Engineering Department at the University of Texas, Austin, TX 78712, USA. Shilpa now works at Bosch Research and Technology Center, 4009 Miranda Ave Suite 150, Palo Alto, CA 94304, USA. {\bf Email:} shilpa.gulati@gmail.com}
\And
Chetan Jhurani\thanks{Tech-X Corporation,	5621 Arapahoe Ave, Boulder, CO 80303, USA. {\bf Email:} chetan.jhurani@gmail.com} \\
\And
Benjamin Kuipers\thanks{Electrical Engineering and Computer Science Department, University of Michigan, Ann Arbor, MI 48109 USA. {\bf Email:} kuipers@umich.edu} \\
}

\title{A Nonlinear Constrained Optimization Framework for Comfortable and Customizable Motion Planning of
Nonholonomic Mobile Robots -- Part II}

\begin{document}

\singlespace
\maketitle

\begin{abstract}
{
In this series of papers, we present a motion planning framework for planning
comfortable and customizable motion of nonholonomic mobile robots such as 
intelligent wheelchairs and autonomous cars.  In Part I, we presented 
the mathematical foundation of our framework, where we model motion discomfort as 
a weighted cost functional and define comfortable motion planning as a 
nonlinear constrained optimization problem of computing trajectories that minimize 
this discomfort given the appropriate boundary conditions and constraints.

In this paper, we discretize the infinite-dimensional optimization problem using 
conforming finite elements.  We describe shape functions 
to handle different kinds of boundary conditions and the choice of unknowns
to obtain a sparse Hessian matrix.  We also describe in detail how
any trajectory computation problem can have infinitely many locally
optimal solutions and our method of handling them.
Additionally, since we have a nonlinear and constrained problem, 
computation of high quality initial guesses is crucial for efficient solution. We
show how to compute them.
}
\end{abstract}

\setcounter{secnumdepth}{2}

%--------------------------------------------------------------------------------------------------
\section{Introduction}
\label{sec:introduction}

In the first paper of this series~\citep{Gulati_2013a}, we formulated motion planning for a nonholonomic mobile robot 
moving on a plane as a constrained nonlinear optimization problem.  This formulation was motivated
by the need for planning trajectories that result in comfortable motion for human users of
autonomous robots such as assistive wheelchairs~\citep{Fehr_2000,Simpson_2008} and autonomous cars~\citep{kpmg_2012}. We reviewed literature from 
robotics and ground vehicles and identified several properties that a trajectory
must have for comfort. We then reviewed motion planning literature in robotics~\citep{Latombe_1991,Choset_2005,LaValle_2006,LaValle_2011a,LaValle_2011b} and concluded that
most existing methods have been developed
for robots that do not transport a human user and have not explicitly addressed issues
of comfort and customization.

We then posed motion planning as a constrained nonlinear optimization problem where
the objective is to compute trajectories that minimize a discomfort cost functional and satisfy 
appropriate boundary conditions and constraints. The discomfort cost functional was formulated
based on comfort studies in road and railway vehicle design. We performed an in-depth analysis 
of conditions under which the cost-functional is mathematically meaningful, a thorough
analysis of boundary conditions, and formulated the constraints necessary for motion comfort
and for obstacle avoidance. We showed that we must be able to impose two
kinds of boundary conditions. In the first kind, the problem is set
in the Sobolev space of functions whose up to second derivatives are square-integrable.
In the second kind, we must allow functions that are singular at
the boundary (with a known strength) but still lie in the same Sobolev
space in the interior. 
 
The above optimization problem is infinite dimensional since it is
posed on infinite dimensional function spaces. This means that we must discretize it as a finite dimensional
problem before it can be solved numerically.  Keeping the problem setting and
requirements mentioned above in mind, it is natural to use the Finite Element Method (FEM)~\citep{Hughes_2000} to
discretize it.  In this paper, we present the details of this discretization and show how to use an 
appropriate finite dimensional subspace
for both kinds of boundary conditions.  We also show the sparsity
structure of the Hessian of the global problem.  Some of the discretized inequality and inequality
constraints, if computed naively, lead to a dense global Hessian.  We avoid this
and keep the global Hessian sparse by introducing auxiliary variables. 

A good initial guess is crucial for solving a nonlinear optimization problem. We
describe a method for computing high quality initial guesses for the optimization problem.

The choice of finite-element method for discretization, the choice of appropriate finite
dimensional subspace for different types of boundary conditions, the choice of appropriate
variables to be solved for, and a method to compute good initial 
guess together result in a fast and reliable solution method.

%--------------------------------------------------------------------------------------------------
\section{Background}
\label{sec:background}
For completeness, we briefly describe the assumptions and problem formulation that are presented
in more detail in Part I of this series. 

\subsection{Motion of a nonholonomic mobile robot moving on a plane} 
\label{sec:motion_of_nonholonomic_robot_on_plane}
We model the robot as rigid body moving on a plane subject to the following nonholonomic constraint
%-----
\begin{equation}
\label{eq:wheeled_robot_constraint}
\dot{x} \sin\theta - \dot{y}\cos\theta = 0.
\end{equation}
%-----
Here dot, $(\dot{\,})$, represents derivative with respect to $t$. 
A motion of such a body can be specified by specifying a travel time $\tau$ and a 
trajectory $\vet{r}$ for $t \in [0,\tau]$. The orientation $\theta(t)$ can be computed
from \Eq{eq:wheeled_robot_constraint}. Essentially, $\theta(t) = \mbox{arctan2}(\dot{\ve{r}}(t))$.
If $\dot{\ve{r}}(t)$ is zero, which means the velocity is zero, then this equation cannot be used.
If the instantaneous velocity is zero at $t = t_0$, and non-zero in a neighborhood
of $t_0$, then $\theta(t_0)$ can be defined as a $\lim_{t \to t_0} \mbox{arctan2}(\dot{\ve{r}}(t))$.

\begin{figure}
\begin{center}
\includegraphics[scale=1,trim = 10 0 0 0]{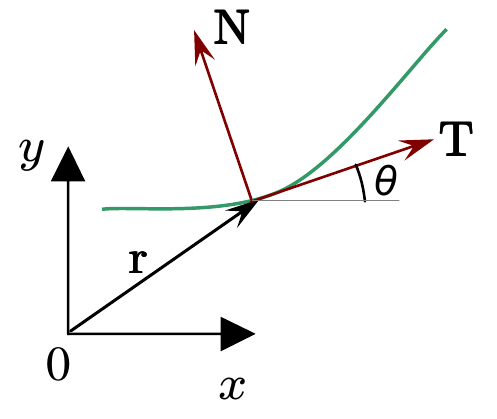}
\end{center}
\caption{Tangent and Normal to a curve}
\label{fig:tangent_and_normal}
\end{figure}

Let the geometric path on which the robot moves be an arc-length parameterized curve $\ve{r}(s)$ 
(see \Fig{fig:tangent_and_normal}). The tangent and normal vectors to the curve are given by
%-----
%
%-----
\begin{equation}
\label{eq:def_tangent_normal}
\begin{split}
\ve{T}(s) &= \frac{ d \widehat{\ve{r}}}{d s}\\
\ve{N}(s) &= \frac{ \frac{d \ve{T}}{d s} }{\norm{ \frac{d \ve{T}}{d s} }}
\end{split}
\end{equation}
The signed curvature $\kappa(s)$ is defined as
\begin{equation}
\label{eq:def_signed curvature}
\kappa(s) = \frac{ d \theta}{d s}
\end{equation}
where $\theta(s)$ is the tangent angle.

\subsection{The discomfort cost functional} 
\label{sec:discomfort_cost_functional}
From review of comfort studies in road and railway vehicles~\citep{Suzuki_1998,Jacobson_1980,Pepler_1980,Forstberg_2000,Chakroborty_2004,Iwnicki_2006}, we concluded that for motion comfort, it is necessary to have  
continuous and bounded acceleration along the tangential and normal directions. It is possible
that the actual values of the bounds on the tangential and normal components are different.
It is also desirable to keep jerk small and bounded.

We model discomfort as the following cost functional $J$:
%-----
\begin{equation}
\label{eq:cost_functional}
J = \tau ~+~ w_{\tny T} \int_{0}^\tau (\dddot{\bf r}\cdot{\bf T})^2 ~dt ~+~ w_{\tny N} \int_{0}^\tau (\dddot{\bf r}\cdot{\bf N})^2 ~dt.
\end{equation}
%-----
Here $\tau$ is the total travel time and ${\bf r}$ is the position of robot at time $t \in [0, \tau]$.
$\dddot{\bf r}$ represents the jerk. $\dddot{\bf r}\cdot{\bf T}$ and $\dddot{\bf r}\cdot{\bf N}$ are
the tangential and normal components of jerk respectively. 
The weights ($w_{\tny T}$ and $w_{\tny N})$ are
non-negative known real numbers. We separate tangential and normal jerk to
allow a choice of different weights ($w_{\tny T}$ and $w_{\tny N}$).

The weights serve two purposes. First, they act as scaling factors for
dimensionally different terms. Second, they determine the relative
importance of the terms and provide a way 
to adjust the robot's performance according to user preferences. For
example, for a wheelchair, some users may not tolerate high jerk and
prefer traveling slowly while others could tolerate relatively high jerks if they
reach their destination quickly. We determine the the typical values of weights 
using dimensional analysis (see Part I~\citep{Gulati_2013a}).

\subsection{Parameterization of the trajectory}
\label{sec:parameterization_of_trajectory}
The trajectory $\ve{r}$ can be parameterized in various ways. We have found that
expressing the trajectory in terms of {\em speed} and {\em orientation} as functions
of a {\em scaled} arc-length parameter leads to relatively simple expressions for all
the remaining physical quantities (such as accelerations and jerks).

Let $u \in [0,1]$.  The trajectory is parameterized by $u$. The starting
point is given by $u = 0$ and the ending point is given by $u = 1$.
Let $\ve{r} = \ve{r}(u)$ denote the position vector of the robot
in the plane.  Let $v = v(u)$ be the speed. Both $\ve{r}$ and $v$ are 
functions of $u$. Let $\lambda$ denote the length of the trajectory. 
Since only the start and end positions are known, $\lambda$ cannot
be specified in advance. It has to be an unknown that will
be found by the optimization process. 

Let $s \in [0, \lambda]$ be the arc-length parameter. 
We choose $u$ to be a scaled arc-length parameter where 
$u = \frac{s}{\lambda}$. The trajectory, 
$\vet{r}, t \in [0, \tau]$ is completely specified by the 
{\em trajectory length} $\lambda$, the {\em speed} $v = v(u)$, 
and the {\em orientation} or the tangent angle $\theta = \theta(u)$ 
to the curve. $\lambda$ is a scalar while speed
and orientation are functions of $u$. These are the three 
unknowns, two functions and one scalar, that will be determined by the optimization process.

With this parameterization, the discomfort cost functional of \Eq{eq:cost_functional} 
can be written as
\begin{equation}
\label{eq:J_total}
J(v, \theta, \lambda) =
\int_0^1 \frac{\lambda}{v} d u
+ w_{\tny T} \int_0^1 \frac{v}{\lambda^3} (v'^2 + v v'' - v^2 \theta'^2)^2 d u
+ w_{\tny N} \int_0^1 \frac{v^3}{\lambda^3} (3 v' \theta' + v \theta'')^2 d u.
\end{equation}
The first integral ($J_\tau$) is the total time, the second integral ($J_{\tny T}$) is
total squared tangential jerk, and the third integral ($J_{\tny N}$) is
total squared normal jerk.

The discomfort $J$ is now a function of the primary unknown functions $v$, $\theta$,
and a scalar $\lambda$, the trajectory length.  All references to time $t$ have
disappeared. Once the unknowns are found via optimization, we can
compute $t$ using the following expression.
%------
\begin{equation}
t = t(u) = \int_0^u \frac{\lambda}{v(u)} du.
\label{eq:t}
\end{equation}
%------

The position vector $\ve{r}(u)$ can be computed via the following integrals.
\begin{equation}
\label{eq:r_by_integral}
\ve{r}(u) = \ve{r}(0) + \lambda \left\{ \int_0^u \cth(u)\, d u, \int_0^u \sth(u)\, d u \right\}.
\end{equation}
If $\theta(u)$ is known, $\ve{r}(u)$ can be computed from \Eq{eq:r_by_integral}.
If $v(u)$  and $\lambda$ are known, $t(u)$ can be computed from \Eq{eq:t}. 
Using these two, we can determine the function $\vet{r}, t \in [0, \tau]$. 

The expressions for tangential acceleration $a_{\tny T}$ and
normal acceleration $a_{\tny N}$ are
\begin{equation}
\label{eq:aT}
a_{\tny T} = \ddot{\ve{r}} \cdot \ve{T} = \frac{v v'}{\lambda}
\end{equation}
\begin{equation}
\label{eq:aN}
a_{\tny N} = \ddot{\ve{r}} \cdot \ve{N} = \frac{v^2 \theta'}{\lambda}.
\end{equation}
Here $\ve{N}$ is the direction normal to the tangent (rotated
$\frac{\pi}{2}$ anti-clockwise).
The signed curvature is given by
\begin{equation}
\label{eq:kappa}
\kappa(u) = \frac {\theta'}{\lambda}
\end{equation}
The angular speed $\omega$ is given by
\begin{equation}
\label{eq:omega}
\omega(u) = \frac{\theta'v}{\lambda}.
\end{equation}

\subsection{Function spaces for $v$ and $\theta$}
\label{sec:function_spaces_for_finite_discomfort}
We now define the function spaces to which $v$ and $\theta$ can belong so that the discomfort
 $J$ in \Eq{eq:J_total} is well-defined (finite).  We have two distinct cases depending on whether 
the speed is zero at an end-point on not.

Let $\Omega = [0,1]$ and $H^2(\Omega)$ be the Sobolev space of functions on $\Omega$
with square-integrable derivatives of up to order 2. Let $f : \Omega \to \mathbb{R}$.
Then
%----
\begin{equation}
\label{eq:H2}
f \in H^2(\Omega)
\stackrel{\rm def}{\iff} \int_{\Omega} \left(\frac{d^j f}{d x^j}\right)^2 d x < \infty
\;\forall\; j = 0,1,2.
\end{equation}
%----

We can show that if $v, \theta \in H^2(\Omega)$, then the integrals of
squared tangential and normal jerk are finite. Using the
Sobolev embedding theorem~\citep{Adams_2003} it can be shown that if
$f \in H^2(\Omega)$, then $f' \in C^0(\Omega)$ and by extension
$f \in C^1(\Omega)$. Here $C^j(\Omega)$ is the space of
functions on $\Omega$ whose up to $j^{th}$ derivatives are
bounded and continuous.  Thus, if $v, \theta \in H^2(\Omega)$,
then all the lower derivatives are bounded and continuous.
Physically this means that quantities like the speed, acceleration, and
curvature are bounded and continuous $-$ all desirable properties
for comfortable motion (\Sec {sec:discomfort_cost_functional}).

We also need that the inverse of $v$ be integrable so that $J_\tau$ is finite.
Inverse of $v$ is integrable if $v$ is
uniformly positive in $[0,1]$.  This is trivially
true if $v$ is uniformly positive, that is, $v \ge \overline{v} > 0$ for some
constant positive $\overline{v}$ throughout the interval $[0, 1]$.  However,
$v$ can be zero at one or both end-points because of the imposed
conditions (see \Sec{sec:boundary_conditions}).

\subsubsection{Positive speed on boundary}
\label{sec:conditions_for_positive_speeds}
Consider the case that $v$ is positive on both end-points.
We make the justifiable assumption
that the trajectory that
actually minimizes discomfort will not have a halt in between.  Thus, if
$v > 0$ on end-points, it remains uniformly positive in the interior
and the discomfort is finite.

\subsubsection{Zero speed on boundary}
\label{sec:conditions_for_zero_speeds}
Consider the case in which $v(0) = 0$.  The case
$v(1) = 0$ can be treated in a similar manner. 
If $v(0) = 0$, $\frac{1}{v}$ must not blow up
faster than $\frac{1}{u^p}$ where $p < 1$ to keep $J_\tau$
finite. Thus, if zero speed boundary conditions are
imposed, we will have to choose $v$ outside $H^2(\Omega)$.
In such a case, at $u=0$, it is sufficient that $v$ approaches zero as $u^p$ where
$\frac{3}{5} < p < 1$ or $p = \frac{1}{2}$.  For the right end point, where $u=1$, replace $u$ with
$(1-u)$ in the condition.  $v \in H^2(\Omega)$ in the interior.

\subsection{Boundary conditions}
\label{sec:boundary_conditions}
The expression for the cost functional $J$ in \Eq{eq:J_total} shows that
the highest derivative order for $v$ and $\theta$ is two.  Thus, for
the boundary value problem to be well-posed we need two boundary
conditions on $v$ and $\theta$ at each end-point $-$ one on the
function and one on the first derivative.

We now relate the mathematical requirement on $v$ and $\theta$ boundary values above
to expressions of physical quantities.
We do this for the starting point only.  The ending point relations
are analogous.

\subsubsection{Positive speed on boundary}
\label{sec:positive_speed_on_boundary}
First, consider the case when $v > 0$ on the starting point.
The speed $v$ needs to be specified, which is quite natural.  The $u$-derivative
of $v$, however, is not tangential acceleration.  The tangential acceleration
is the $t$-derivative and is given by \Eq{eq:aT}. It is $\frac{v v'}{\lambda}$.  Here $v$ is known
but $\lambda$ is not. Thus specifying
tangential acceleration gives us a constraint equation and not directly
a value for $v'(0)$.  This is imposed as an equality constraint.
Similarly, fixing a value for $\theta$ on starting point is natural.
We ``fix'' the values of $\theta'(0)$ by fixing the signed curvature $\kappa = \frac{\theta'}{\lambda}$.
As before, this leads to an equality constraint relating
$\theta'(0)$ and $\lambda$ if $\kappa \neq 0$.  Since choosing a meaningful non-zero
value of $\kappa$ is difficult, it is natural to impose $\kappa = 0$.  In this
case $\theta'(0) = 0$ can be imposed easily.

\subsubsection{Zero speed on boundary}
\label{sec:zero_speed_on_boundary}
If $v(0) = 0$, then, as seen in \Sec{sec:function_spaces_for_finite_discomfort},
$v(u)$ must behave like $u^p$ for $\frac{3}{5} < p < 1$ or $p = \frac{1}{2}$ near $u=0$ and
$v'(u) \sim u^q$ for $-\frac{2}{5} < q < 0$ or $q = -\frac{1}{2}$ respectively.  
In this case
\begin{equation}
\label{eq:vdash_strength}
v'(u) \sim u^{-1/3}
\end{equation}
is the appropriate strength of the singularity for $v'(u)$ at the starting point. 

\subsection{Obstacle avoidance constraints}
\label{sec:obstacle_avoidance_constraints}
We model the ``forbidden'' region formed by the obstacles as a union
of star-shaped domains with boundaries that are closed curves with piecewise continuous second derivative.
A set in $\mathbb{R}^n$ is called a star-shaped domain if there exists at least one point
${\bf x}_0$ in the set such that the line segment connecting ${\bf x}_0$ and ${\bf x}$ lies in the set
for all ${\bf x}$ in the set. Intuitively this means that there exists at least one point
in the set from which all other points are ``visible''. We will refer to such a point 
${\bf x}_0$ as a \emph {center} of the star-shaped domain.

\begin{figure}
\begin{center}
\includegraphics[scale=0.6]{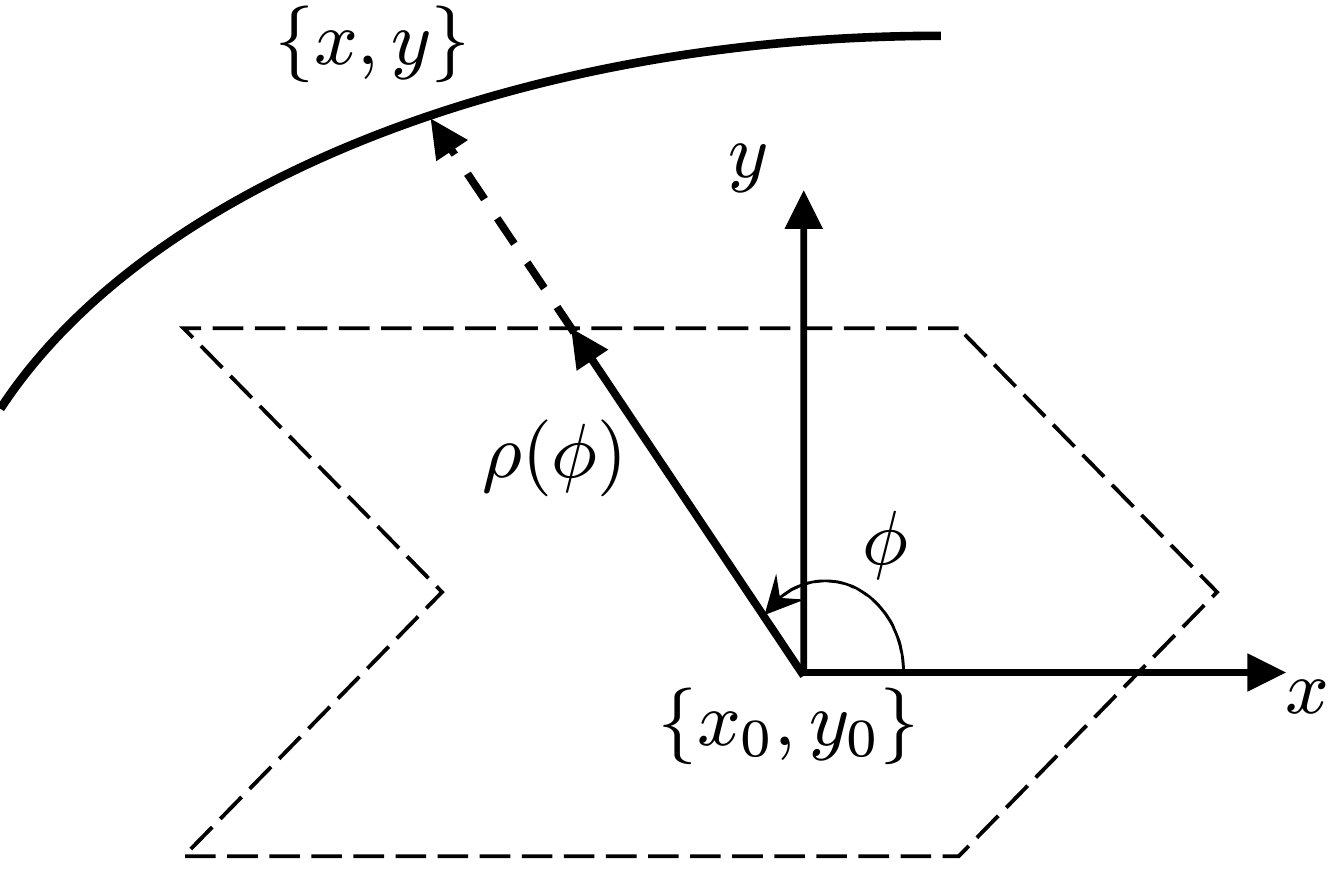}
\end{center}
\caption
{Notation for star-shaped obstacles.}{\small{A non-convex star-shaped obstacle is shown with
its ``center'' $\{ x_0, y_0 \}$ and a distance function $\rho = \rho(\phi)$.
The distance function gives a single point on the boundary for $\phi \in [0,2\pi]$.
The robot trajectory must lie outside the obstacle.}}
\label{fig:star_shaped}
\end{figure}

Let an obstacle be specified by its boundary in polar coordinates
that are centered at $\ve{r}_0 = \{x_0, y_0\}$.  Each $\phi \in [0, 2\pi)$ gives a
point on the boundary using the distance $\rho(\phi)$ from the
obstacle origin. The distance function
$\rho$ must be periodic with a period $2 \pi$. See \Fig{fig:star_shaped}.

Suppose we want a point $\ve{r} = \{x, y\}$ to be outside the obstacle boundary.
Define $C(\ve{r})$ as
%----
\begin{equation}
\label{eq:star_shaped}
C(\ve{r}) = \norm{\ve{r} - \ve{r}_0}_2 - \rho( \mbox{arctan2}(\ve{r} - \ve{r}_0))
\end{equation}
%----
where the subscript 2 refers to the Euclidean norm. $C(\ve{r}) \ge 0 \iff$ the point
$\ve{r}$ is outside the obstacle.
%----
%----
The constraint function $C(\ve{r})$ is piecewise differentiable for all
$\ve{r}$ except at a single point $\ve{r} = \ve{r}_0$.  If $\ve{r} = \ve{r}_0$
by chance, which is easily detectable, we know that the $\ve{r}$ is inside the obstacle and can
perturbed to avoid this undefined behavior. Note that  $C(\ve{r})$ remains bounded inside
the obstacle.

%----------------------------------------------------------------------------------------------------
\section{The infinite-dimensional optimization problem}
\label{sec:infinite_dimensional_optimization_problem}
We now summarize the nonlinear and constrained trajectory optimization problem
taking into account all input parameters, all the boundary conditions,
and all the constraints. This is the ``functional'' form of the problem (posed in function
spaces).  We will present an appropriate discretization procedure valid for
all input combinations in the \Sec{sec:finite_element_discretization}.

Minimize the discomfort functional $J$, where
$$
J(v, \theta, \lambda) =
\int_0^1 \frac{\lambda}{v} d u
+ w_{\tny T} \int_0^1 \frac{v}{\lambda^3} (v'^2 + v v'' - v^2 \theta'^2)^2 d u
+ w_{\tny N} \int_0^1 \frac{v^3}{\lambda^3} (3 v' \theta' + v \theta'')^2 d u,
$$

given the following boundary conditions for both starting point and ending point
\begin{itemize}
	\item position ($\ve{r}_0$, $\ve{r}_\tau$),
	\item orientation ($\theta_0$, $\theta_\tau$),
	\item signed curvature ($\kappa_0$, $\kappa_\tau$),
	\item speed ($v_0 \ge 0$, $v_\tau \ge 0$),
	\item tangential acceleration ($a_{{\tny T},0}$, $a_{{\tny T},\tau}$),
\end{itemize}
and constraints on allowable range of
\begin{itemize}
	\item speed ($v_{min} = 0, v_{max}$),
	\item tangential acceleration ($a_{{\tny T},min}, a_{{\tny T},max}$),
	\item normal acceleration ($a_{{\tny N},min}, a_{{\tny N},max}$),
	\item angular speed ($\omega_{min}, \omega_{max}$),
	\item curvature, if necessary ($\kappa_{min} = 0, \kappa_{max}$),
\end{itemize}
and
\begin{itemize}
\item number of obstacles $N_{obs}$, 
\item locations of obstacles $\{ \ve{c}_i \}_{i=1}^{N_{obs}}$ 
\item representation of obstacles that allows computation of $\{ \rho_i(\phi)\}_{i=1}^{N_{obs}}$, for $\phi \in [0,2\pi)$ 
\end{itemize}
and
\begin{itemize}
\item an initial guess for $(v(u), \theta(u), \lambda)$, in $u \in [0,1]$, 
\item weights $w_{\tny T} > 0$ and $w_{\tny N} > 0$.
\end{itemize}

The constraint on starting and ending position requires that
\begin{equation}
\label{eq:r_by_integral_constr}
\ve{r}_\tau - \ve{r}_0 = \lambda \left\{ \int_0^1 \cth\, d u, \int_0^1 \sth\, d u \right\}.
\end{equation}

Staying outside all obstacles requires that
$$
\norm{\ve{r}(u) - \ve{c}_i}_2 - \rho_i( \mbox{arctan2}(\ve{r}(u) - \ve{c}_i)) \ge 0
\;\forall \; i \in 1, \ldots, N_{obs}, \mbox{ and} \; \forall \; u \in [0,1]
$$
where
$$
\ve{r}(u) = \ve{r}(0) + \lambda \left\{ \int_0^u \cth\, d u, \int_0^u \sth\, d u \right\}.
$$
As a post-processing step, we compute time $t$ as a function of $u$ using
$$
t = t(u) = \int_0^u \frac{\lambda}{v(u)} du
$$
and convert all quantities ($v, \theta, \ve{r},$ and their derivatives) from
$u$ domain to $t$ domain.

%--------------------------------------------------------------------------------------------------
\section{A finite element discretization of the infinite-dimensional optimization problem} 
\label{sec:finite_element_discretization}

We first discuss the case when there is no singularity in the speed $v$.
This is the case when the given boundary speeds are positive.  In this case, $v \in H^2(\Omega)$
as described in \Sec{sec:function_spaces_for_finite_discomfort}, where $\Omega = [0,1]$.
We assume that $\theta \in H^2(\Omega)$ always (whether $v$ is singular or not) because it is sufficient for
the discomfort to be finite.  Thus, to discretize the problem,
it is natural to use the basis functions in $C^1(\Omega)$, the space of
functions that are continuous and have continuous first derivatives.  This
makes the second derivative of $v$ and $\theta$ discontinuous but its
square is still integrable.

\subsection{Basis functions}
\label{sec:basis_functions}

We minimize the discomfort and satisfy all the constraints in a finite dimensional subspace
of $C^1(\Omega)$.  We make the following choice.
\begin{eqnarray}
\label{eq:finite_dimensional_v_theta}
v^h(u) &=& \sum_{i=1}^q \alpha_i^v \chi_i(u)\\
\theta^h(u) &=& \sum_{i=1}^q \alpha_i^\theta \chi_i(u)
\end{eqnarray}
Here $\chi_i(u) \in C^1(\Omega)$ are basis functions for this problem.
The symbol $h$ traditionally denotes a measure of the ``mesh width''
to distinguish the approximate solution from the ``exact''
infinite dimensional solution.
The unknown scalar values $\alpha_i^v$ and $\alpha_i^\theta$ for $i = 1,\ldots,q$
are the degrees of freedom (DOFs) which are to be determined via ``solving'' the optimization problem
of \Sec{sec:infinite_dimensional_optimization_problem}.  For reasonable choices of $\chi_i(u)$,
as $q$ increases the finite dimensional solution approaches the
exact solution.

Usually, one chooses $\chi_i(u) \in \Omega$ that are easy and cheap to compute,
and have local support.  Piecewise polynomial functions that have sufficient
differentiability are good candidates.  By having local support, we mean
the functions are non-zero only over a limited interval and not on whole $\Omega$.
This has two main advantages.  First, while performing integration to compute $J$,
the product of $\chi_i$ and $\chi_j$ is zero over most of the interval.
This leads to $O(n)$ rather than $O(n^2)$ interactions.  Second, the global
Hessian matrix is sparse rather than being dense.  Typically for 1D problems,
the matrix has a small constant band-width.

For our problem, we first divide the interval $[0,1]$ into $n$ equal-size
intervals of length $h = \frac{1}{n}$. Each of these intervals is an {\em element}.  
Thus we have $n$ elements and $n + 1$ equidistant points or {\em nodes}.  The collection
of elements and nodes is the finite element {\em mesh}.

\begin{figure}
{
\centering
\includegraphics[scale=0.65]{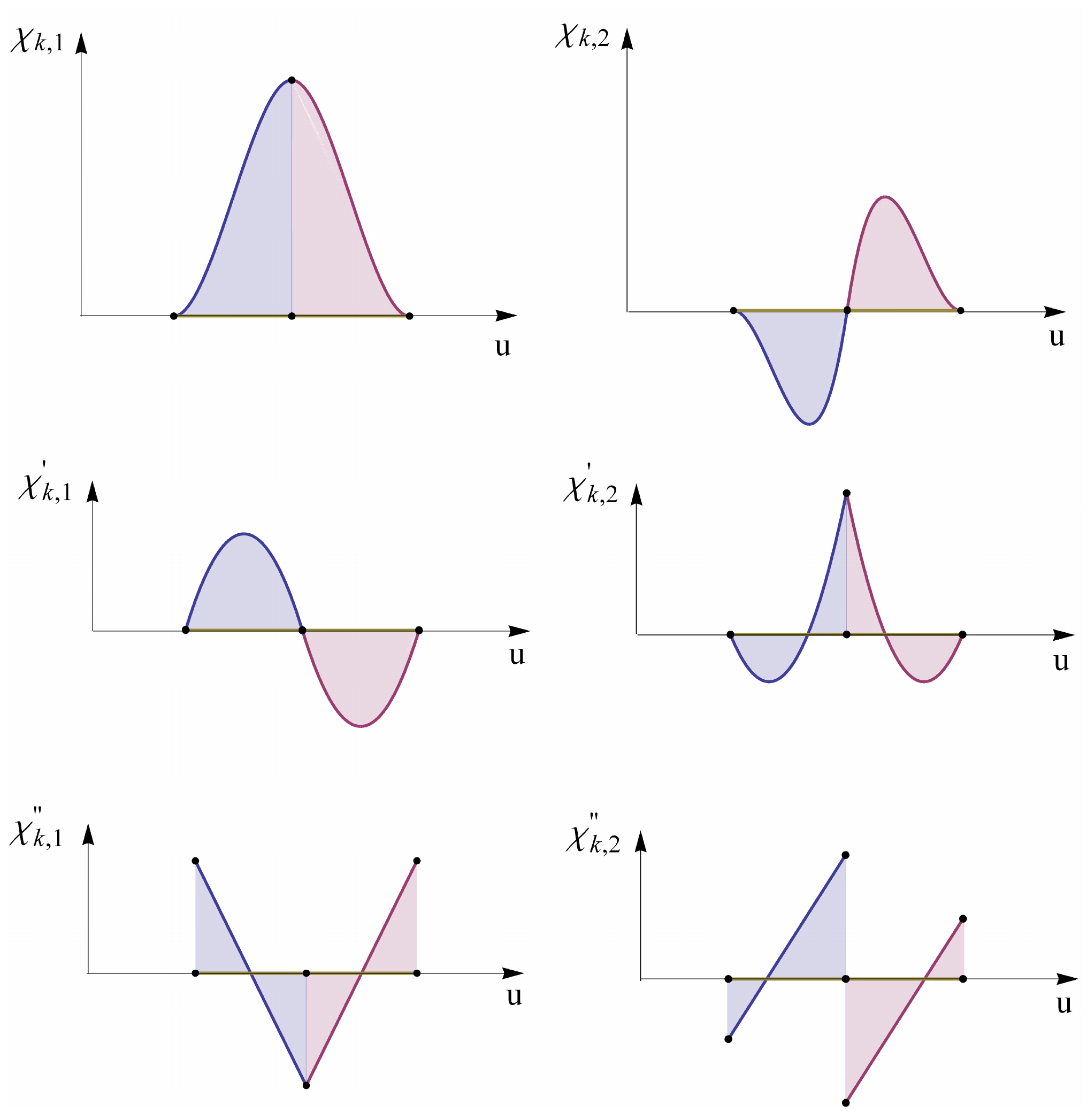}
}
\caption
{Cubic Hermite basis functions and their first and second derivatives.} 
{\small Two kinds of basis functions are
defined at each node such that each is zero everywhere except on the two
elements sharing the node.  The first kind, $\chi_{k,1}$ has
a value of 1 and a zero derivative at the associated node $k$. Its
value and derivatives are zero on both adjacent nodes. The second kind, $\chi_{k,2}$ has
a value of zero and a unit derivative at the associated node $k$. Its
value and derivatives are zero on both adjacent nodes. Both $\chi_{k,1}$
and $\chi_{k,2}$ are square integrable on $u \in [0,1]$.}
\label{fig:c1_basis}
\end{figure}

\begin{figure}
{
\centering
\includegraphics[scale=0.75]{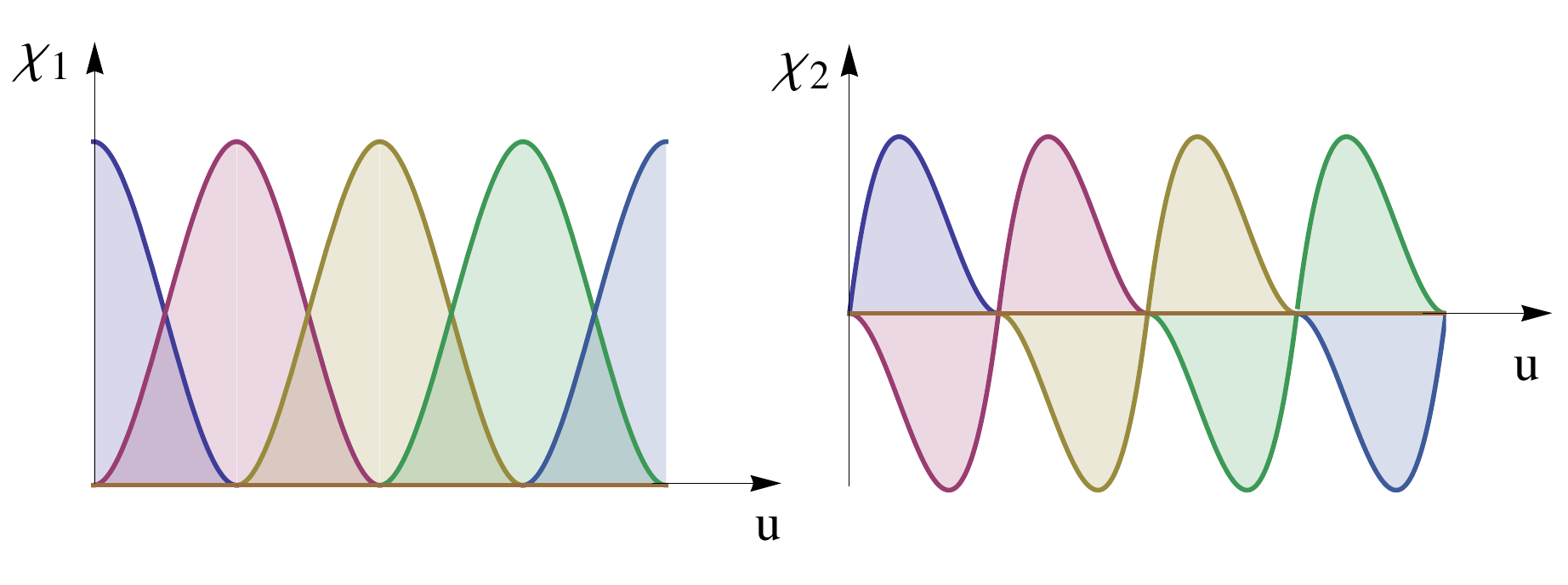}
}
\caption
{Cubic Hermite basis functions on five consecutive nodes.} 
{\small The two kinds of basis functions on any node are translations of the respective 
basis function of \Fig{fig:c1_basis}. The
basis functions on a boundary node are truncated.}
\label{fig:c1_basis_multiple} 
\end{figure}

At each node we define two piecewise polynomial functions that
are non-zero only on the two elements surrounding the node.  This is the
standard cubic Hermite basis for problems posed in the $H^2$ space in 1D.
See \Fig{fig:c1_basis} for both the functions and their first and
second derivatives.  The first kind of basis function is 1 on the
node with which it is associated and has zero derivative there.
The second function is zero on the corresponding node and has unit derivative there.
Additionally, both are zero with first derivatives also zero on two surrounding
nodes.  Thus, in all, we have $2(n+1)$ basis functions and unknown scalars
each for $v$ and $\theta$.  Because of the above-mentioned properties
each basis functions belongs to $C^1(\Omega) \subset H^2(\Omega)$. Note that
the basis functions on the boundary points are slightly different.  They
are not extended outside the interval.  \Fig{fig:c1_basis_multiple} shows
5 basis functions of each kind for $n=4$.  As seen, the boundary basis
functions are truncated. It would not matter anyway what their values outside
the interval are since no integration is performed outside.  The other
basis functions are translations of each other.

\subsection{Element shape functions}
\label{sec:element_shape_functions}
In FEM practice, we define basis functions in terms of ``shape functions''.
The shape functions are defined only on a single reference element
and multiple shape functions placed on neighboring elements are joined
to create a single basis function.  This requires that shape functions
have appropriate values and derivative on reference element boundary
so that the basis functions are valid for the problem.

\begin{figure}[t!]
\begin{center}
\includegraphics[scale=0.6]{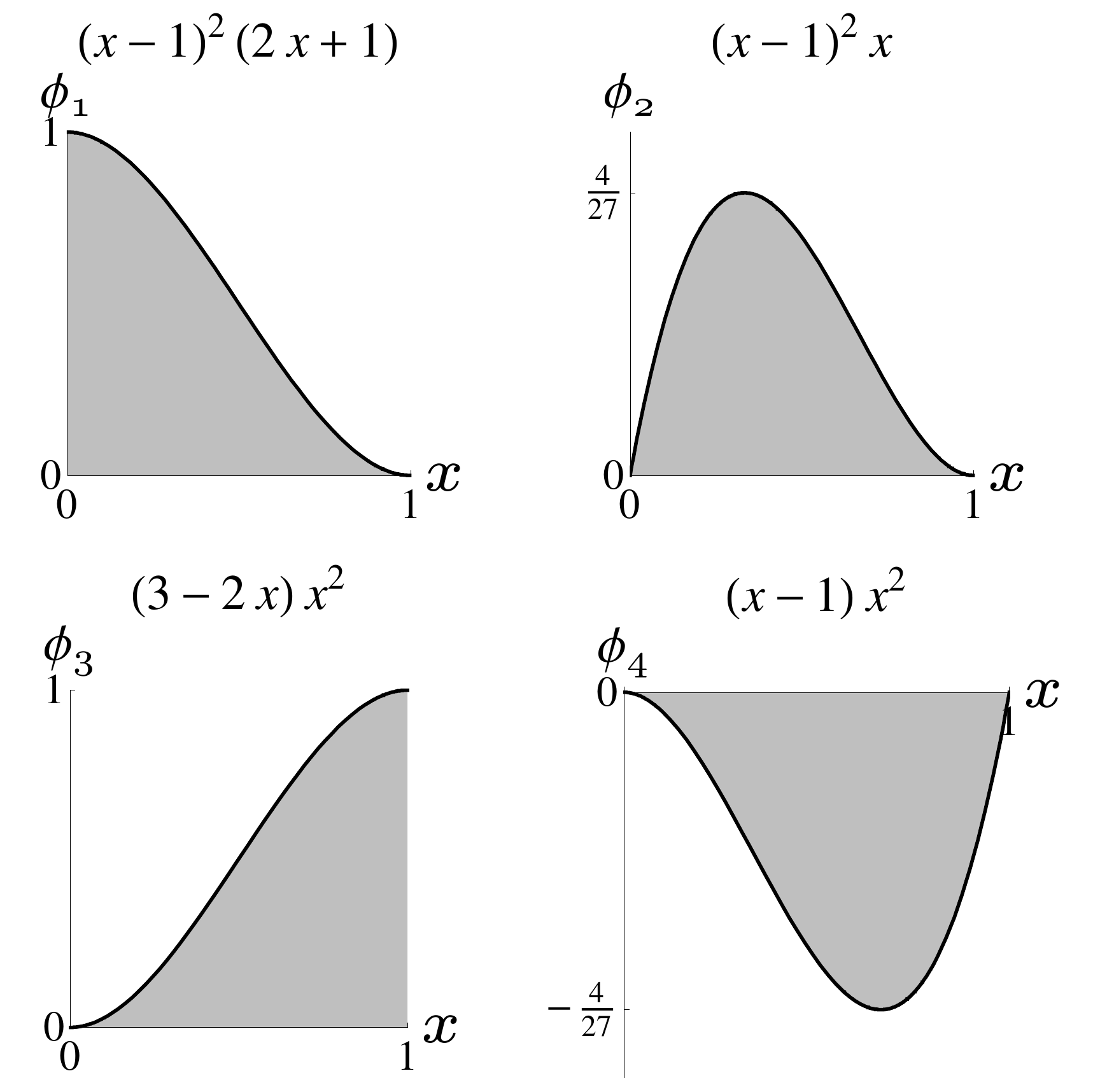}
\end{center}
\caption{Cubic Hermite shape functions on a reference element.} {\small Four shape
functions are defined on each element. Each is a cubic polynomial on $x = [0,1]$ 
where  $x$ is the local coordinate on the element.}
\label{fig:c1_shape}
\end{figure}

\Fig{fig:c1_shape} shows the four cubic Hermite shape functions on a single
reference element $[0,1]$.  The shape functions are
\begin{eqnarray*}
\phi_1(x) &=& (x-1)^2 (1 + 2 x)\\
\phi_2(x) &=& (x-1)^2 x\\
\phi_3(x) &=& (3-2 x) x^2\\
\phi_4(x) &=& (x-1) x^2
\end{eqnarray*}
For maintaining basis function continuity, each $\phi_i$ satisfies
$\phi_i(0) = \phi_i'(0) = \phi_i(1) = \phi_i'(1) = 0$, except
$\phi_1(0) = \phi_2'(0) = \phi_3(1) = \phi_4'(1) = 1$.

\subsection{Singular shape functions at boundary}
\label{sec:singular_elems}

For the case when either one or both the boundary points have
zero speed specified, $v(u)$ is singular at the corresponding
boundary with $v'(u)$ infinite with singularity $u^{-1/3}$
when acceleration is also zero and $u^{-1/2}$ if acceleration
is non-zero.
Thus, $v$ as a function does not belong to $H^2(\Omega)$.
However, $v$ does belong to $H^2$ in the interior as shown
in \Sec{sec:function_spaces_for_finite_discomfort}.

After a FEM mesh is decided, we take this into account and
do not use the above-mentioned regular shape functions for $v$
on the element(s) near the boundary with zero speed.
For the interior elements, however, no change is done
and the regular shape functions are used.

We now derive the new singular shape functions for the left boundary $(u \in [0,h])$
element and the singular shape functions for the right boundary
element can be derived using symmetry.

\begin{figure}
\begin{center}
\includegraphics[scale=0.65]{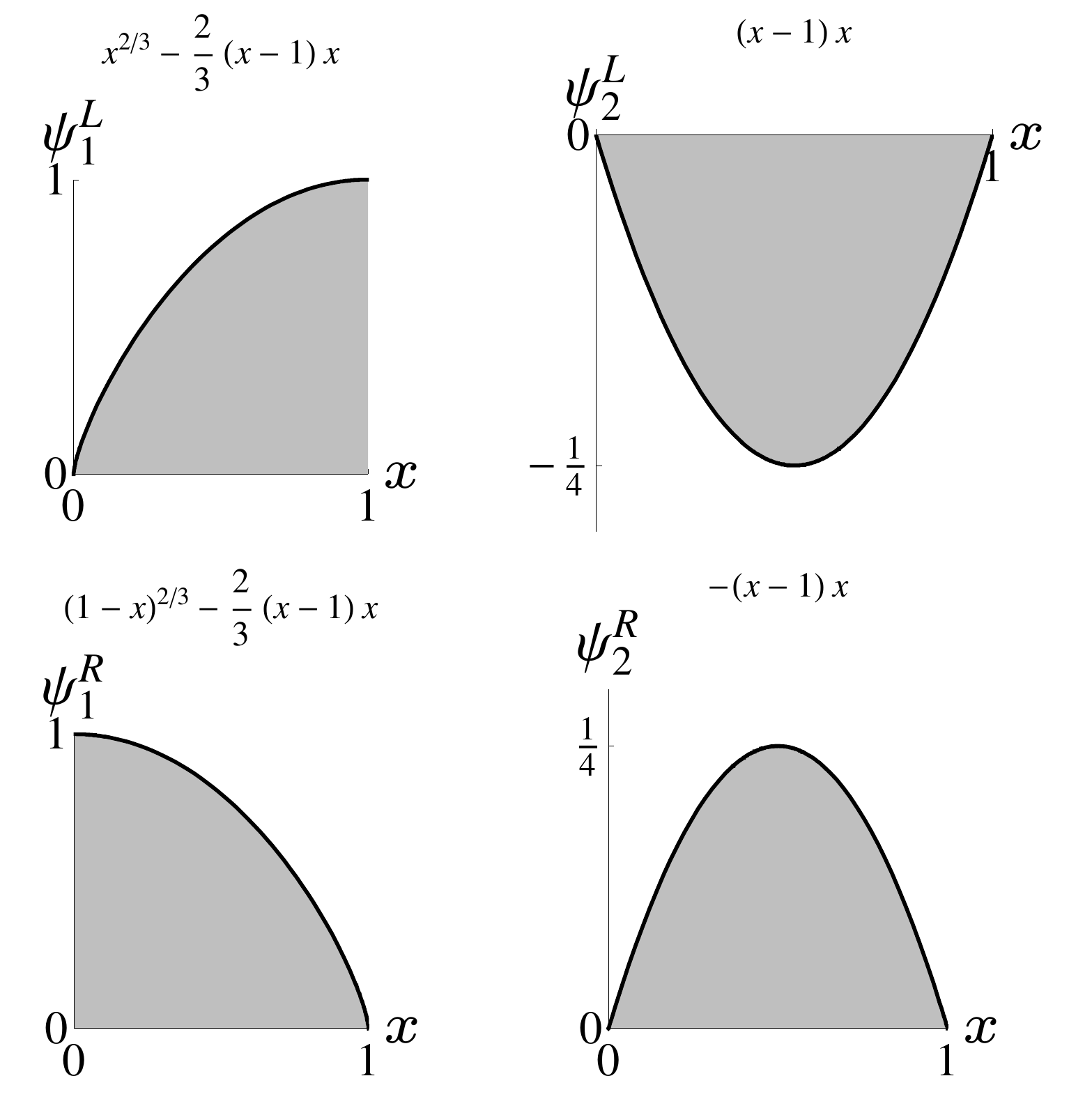}
\end{center}
\caption
{Singular shape functions on boundary elements for zero speed and zero acceleration boundary conditions.}{\small  Two singular shape functions are defined on a boundary
element. Consider zero speed condition on left element. The first shape
function $\psi^L_1$ has a value
of 1 and zero derivative on the right to match the shape function $\phi_1$ of \Fig{fig:c1_shape} on the next
element. The second shape
function $\psi^L_2$ has a value of 0
and a unit derivative on the right to match the shape function $\phi_2$ of \Fig{fig:c1_shape} on the next
element. Both $\psi^L_1$ and $\psi^L_2$ have a  value of zero on the left because speed is zero. The singular shape functions for zero speed boundary conditions on right are similarly defined.}
\label{fig:singular_shape_func}
\end{figure}

\begin{figure}[t!]
\begin{center}
\includegraphics[scale=0.65]{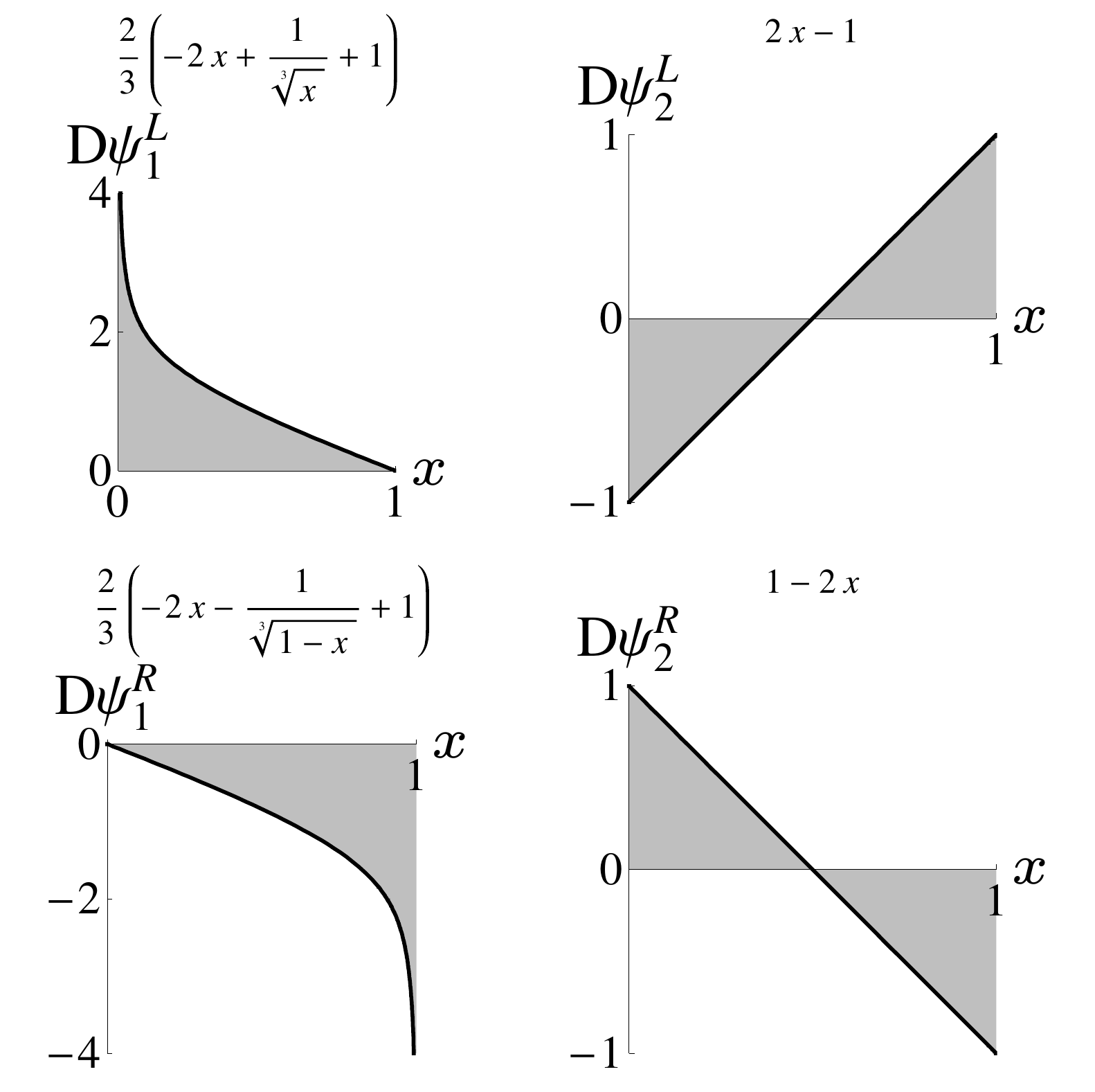}
\end{center}
\caption
{\small First and second derivatives of singular shape functions for zero speed and zero acceleration boundary conditions.} {$\psi_1^L$ and $\psi_1^R$
tend to infinity as $x^{-\frac{2}{3}}$ as they approach the boundary. 
$\psi_2^L$ and $\psi_2^R$ are smooth.}
\label{fig:singular_shape_dfunc}
\end{figure}

We need at least two shape functions so that the two shape functions coming from the $[h,2h]$ element
can be matched.  Denote them by $\psi_1^L$ and $\psi_2^L$.
The function value and the function derivative both must be matched at $h$.
For a reference element shape function, this means $\psi_1^L(1) = 1, {\psi_1^L}'(1) = 0,
\psi_2^L(1) = 0, {\psi_2^L}'(1) = 1$.
Both $\psi_1^L$ and $\psi_2^L$ must be zero on $u=0$ because the speed is zero there.
Thus, the Dirichlet boundary condition is imposed explicitly.
Finally, as $x \to 0$, one of the functions must behave as $x^{p}$, for
$p = \frac{2}{3}$ or $p = \frac{1}{2}$, to match
the singularity in \Eq{eq:vdash_strength}. It can be seen that the choice
\begin{eqnarray*}
\psi^L_1(x) &=& x^{p}+ p (1-x) x\\
\psi^L_2(x) &=& (x-1) x
\end{eqnarray*}
satisfies all these requirements.  \Fig{fig:singular_shape_func} shows the
four shape functions, two for the left element and two for the right element
for the case $p = \frac{2}{3}$.
\Fig{fig:singular_shape_dfunc} shows that $\psi_1^L$ and $\psi_1^R$
are singular when approaching the boundary (shown for $p = \frac{2}{3}$ only).
The other two functions $\psi_2^L$ and $\psi_2^R$ are smooth.

\subsection{The finite-dimensional optimization problem}
\label{sec:finite_dim_opt}
With the choice of basis functions described above, we
can express $v^h(u)$ and $\theta^h(u)$ as:
\begin{eqnarray}
\label{eq:finite_dimensional_v_theta_2}
v^h(u) &=& \sum_{i=1}^{n+1} v_i \chi_{i,1}(u) + \sum_{i=1}^{n+1} v'_i \chi_{i,2}(u)\\
\theta^h(u) &=& \sum_{i=1}^{n+1} \theta_i \chi_{i,1}(u) + \sum_{i=1}^{n+1} \theta'_i \chi_{i,2}(u)
\end{eqnarray}
where $v_i$, $v'_i$, $\theta_i$, and $\theta'_i$ for $i = 1, \ldots, n$ are the unknown nodal
values and $\chi_{i,1}(u)$ and $\chi_{i,2}(u)$ are the two kinds of basis functions described in
the previous section.

For optimization, the values of cost, its gradient and Hessian, the values of constraints,
and the gradient and Hessian of each constraint are required. For efficiency,
it is desirable that cost and constraint Hessians be sparse. We will see later
that for the Hessian of obstacle avoidance constraints to be sparse, it is useful
to introduce $2N$ additional unknowns in the form of position $\ve{r}_i = \left\{x_i, y_i\right\}_{i=1}^{N}$
at $N$ points. 

Thus, our objective now is to determine the values of these unknowns and the
unknown path length $\lambda$ that
minimize the cost functional and satisfy the boundary
conditions and constraints described in (\Sec{sec:infinite_dimensional_optimization_problem}).

\subsubsection{Numerical integration for computing the integrals in the cost functional and constraints}
\label{sec:gauss}

We use Gauss quadrature formulas to compute the integrals in the
cost functional and constraints.  When using $m$ integration
points in an interval, the formulas are accurate for polynomials
of degree up to $2m - 1$.  For the regular $C^1$ basis functions,
which have maximum polynomial degree 3, it is easily seen that
the square tangential jerk is a polynomial of degree 23,
and the squared normal jerk is a polynomial of degree 17.
See \Eq{eq:J_total}. These are polynomials in $u$ and not $t$.
Hence, 12 Gauss points will give exact integrals up to floating
point accuracy.  Of course, the integrands being polynomials,
the integrals corresponding to $J_T$ and $J_N$ can be evaluated
without using Gauss points (if one is ready to work with
complex algebraic expressions). But the other integrals, for
$J_\tau$ and those relating $\ve{r}$ to $\theta$ (\Eq{eq:r_by_integral}),
must be evaluated numerically.  Hence, we use 12 Gauss points to evaluate
all integrals.

\subsection{Imposing constraints}

We need to impose multiple equality and inequality constraints
while minimizing the cost functional. Some of the equality constraints affect a single DOF each and hence
they can be used to eliminate the particular unknown.  The others relate multiple
DOFs and must be imposed as an equality explicitly. The equality constraints are
described  below.
\begin{itemize}
	\item Fix end-point positions ($\ve{r}_0$, $\ve{r}_\tau$) by eliminating
	the unknowns $x$ and $y$ on the first and last nodes.
	\item Fix end-point orientations ($\theta_0$, $\theta_\tau$) by eliminating
	the unknown $\theta$ on the first and last nodes.
	\item Relate start and end position $\ve{r}_\tau - \ve{r}_0 = \lambda \left\{ \int_0^1 \cth\, d u, \int_0^1 \sth\, d u \right\}$ (\Eq{eq:r_by_integral_constr}) by computing the integrals as described
	in \Sec{sec:obstacle_constr}, \Eq{eq:r_adjacent}.
	%$\int_0^1 \cth\, d u$ and $\int_0^1 \sth\, d u$.
	\item Fix end-point speeds ($v_0 \ge 0$, $v_\tau \ge 0$) by eliminating
	the unknown $v$ on the first and last nodes.
	\item If speed on an end-point is positive, tangential acceleration $a_{\tny T}$ must be
	specified at that end-point. Impose $\frac{vv'}{\lambda} = a_{\tny T}$ on that
	end point. Otherwise, this constraint will be automatically imposed by using
	singular shape functions.
	\item Impose specified end-point curvature $\kappa$ by imposing $\theta' = \lambda \kappa$ 
	on each end-point.
\end{itemize}

The inequality constraints that are not related to obstacles avoidance are as follows.
We must maintain
\begin{itemize}
	\item velocity in $[v_{min} = 0, v_{max}]$,
	\item tangential acceleration in $[a_{T,min}, a_{T,max}]$,
	\item normal acceleration in $[a_{N,min}, a_{N,max}]$, and
	\item angular velocity in $[\omega_{min}, \omega_{max}]$.
	\item curvature in $[\kappa_{min} = 0, \kappa_{max}]$,
\end{itemize}

Note that these must be maintained for each $u \in [0,1]$ in the infinite dimensional
optimization problem.  For the discretized version, we choose the Gauss
integration points and impose that these quantities remain in the specified
range {\em only} on those points.  Thus, for a mesh with $n$ elements,
each inequality above results in $12 n$ constraints (assuming 12 points are
used as discussed in \Sec{sec:gauss}).  The values of these physical quantities
on each Gauss point is a function of the DOFs on element nodes.
Thus, these constraints are local.  They are not affected when DOFs of non-element nodes change.

There are two important reasons to keep the values within range on Gauss points
as opposed to on some other, say, uniform set of points.  First, since we
use $v$ at the Gauss point to compute the integrals, it is more important that
$v$ remain non-negative there to avoid problems of large negative values of $J$.
Second, since $v$ and $\theta$ are already computed there it saves extra computation.

\subsection{Obstacle avoidance constraints}
\label{sec:obstacle_constr}
Staying outside obstacles, if present, requires additional inequality constraints.  For this we
pick $N$ uniformly separated points in the interval $[0,1]$ and impose
the constraint that each of $\ve{r}$ on the $N$ points remain outside each
of $N_{obs}$ obstacles.  This leads to $N \times N_{obs}$ constraints. In our implementation
we make $N = n M + (n+1)$, so that if the distribution is uniform, each
element has $M$ such points in the interior and each node is a point too.
Two of these $N$ points are the boundary points which must be outside all
obstacles for the optimization problem to have a feasible solution.
If the robot boundary is not circular and we choose $P$ points on
the robot's boundary, then the number of constraints is $N \times N_{obs} \times P$.
%This is an exhaustive approach and we discuss how some of these constraints
%may be removed intelligently in \Sec{}.

We come back to obstacle related constraint relating a single obstacle and
a single point on the trajectory.  One could simply relate the position
at the point with $\theta(u)$ and $\lambda$ using \Eq{eq:r_by_integral}, and
use \Eq{eq:star_shaped} to impose conditions on $\theta(u)$ and $\lambda$.
However, because of the structure of \Eq{eq:star_shaped} and because
$\ve{r}(u)$ depends on {\em all} $\theta$ DOFs of nodes that
are before $u$, the Hessian of this constraint is not sparse.  This would
lead to efficiency problems when doing iterations in the numerical optimization
process. Even computing the dense Hessian would be very costly as $N_{obs}$ and
$N$ increase. We must work around this elimination approach of imposing the obstacle
related constraints.

To avoid the dense Hessian of obstacle constraint, we make two changes to
the simplistic approach.  First, we do not use \Eq{eq:r_by_integral_constr}
for eliminating $\ve{r}$ but keep $\ve{r}$ as an unknown function.
Second, we relate adjacent $\ve{r}$'s via \Eq{eq:r_by_integral} as follows.
\begin{equation}
\label{eq:r_adjacent}
\ve{r}(u_j) - \ve{r}(u_{j-1}) = \lambda \left\{ \int_{u_{j-1}}^{u_j} \cth\, d u, \int_{u_{j-1}}^{u_j} \sth\, d u \right\}.
\end{equation}
Here $j$ goes from 1 to $N-1$.
What this change does is that, as long as adjacent $\ve{r}(u_j)$ and $\ve{r}(u_{j-1})$ belong to a maximum
two adjacent elements, the equation above relates only a small number of
local DOFs.  Secondly, since each $\ve{r}(u_j)$ is now a legitimate
unknown, it can be used to impose the inequality constraint \Eq{eq:star_shaped}
without $\theta$ being involved.

This new approach does have a price, however.  We have increased the number of
unknowns and hence increased the size of the gradient vector and Hessian matrix.
But this is a small price to pay considering that the sparsity is still maintained,
the amount of computation does not grow, and equations are local in nature.
We explore the sparsity pattern more in \Sec{sec:global_derivs} ahead.

\subsection{Element and global gradient and hessian}
\label{sec:global_derivs}

An important choice in the FEM discretization of any variational problem
is the ordering of all the unknowns when forming the global Hessian
matrix.  A good choice simplifies the assembly process as well
as could lead to useful structural sparsity.

We have four kinds of DOFs.  For simplicity, we discuss the regular case
and where boundary conditions are not yet imposed.  The singular case
differs in minor details only that does not affect the ordering process.
The four kinds are as follows.
\begin{itemize}
	\item four unknowns each on $n + 1$ node $- v, v', \theta, \theta'$
	\item $N$ $x$ and $N$ $y$ unknowns
	\item a scalar unknown $\lambda$
\end{itemize}
The unknowns are ordered in the same sequence shown above starting from $u=0$
and going to $u=1$.

The ordering chosen above means that each DOF except $\lambda$ interacts
with DOFs on two elements. The scaled arc-length
parameter, $\lambda$, is global by its nature and interacts with all other
DOFs.  Hence, the global Hessian matrix is sparse.  Some
interactions lead to linear equations, so they do not affect the
Hessian.   This is the case for $x$, $y$, and $\lambda$ interactions in \Eq{eq:r_adjacent}.

\begin{figure}
\begin{center}
\subfigure[A finite element mesh with 3 elements along with $N = 10$ $\left\{x,y\right\}$ pairs for obstacle avoidance.]{\label{fig:global_divisions}
\includegraphics[scale=0.48]{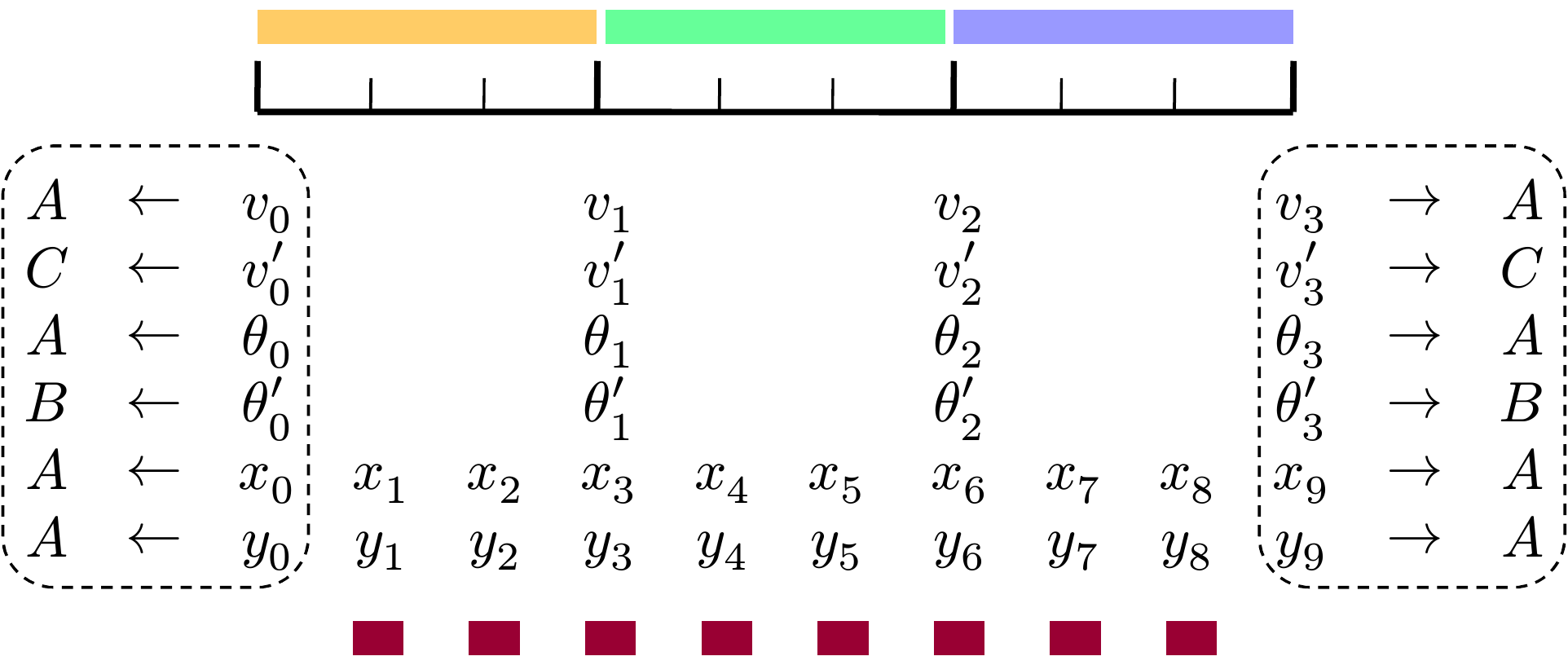}}\\[6pt]
\subfigure[Connectivity structure]{\label{fig:global_matrix}
\includegraphics[scale=0.48]{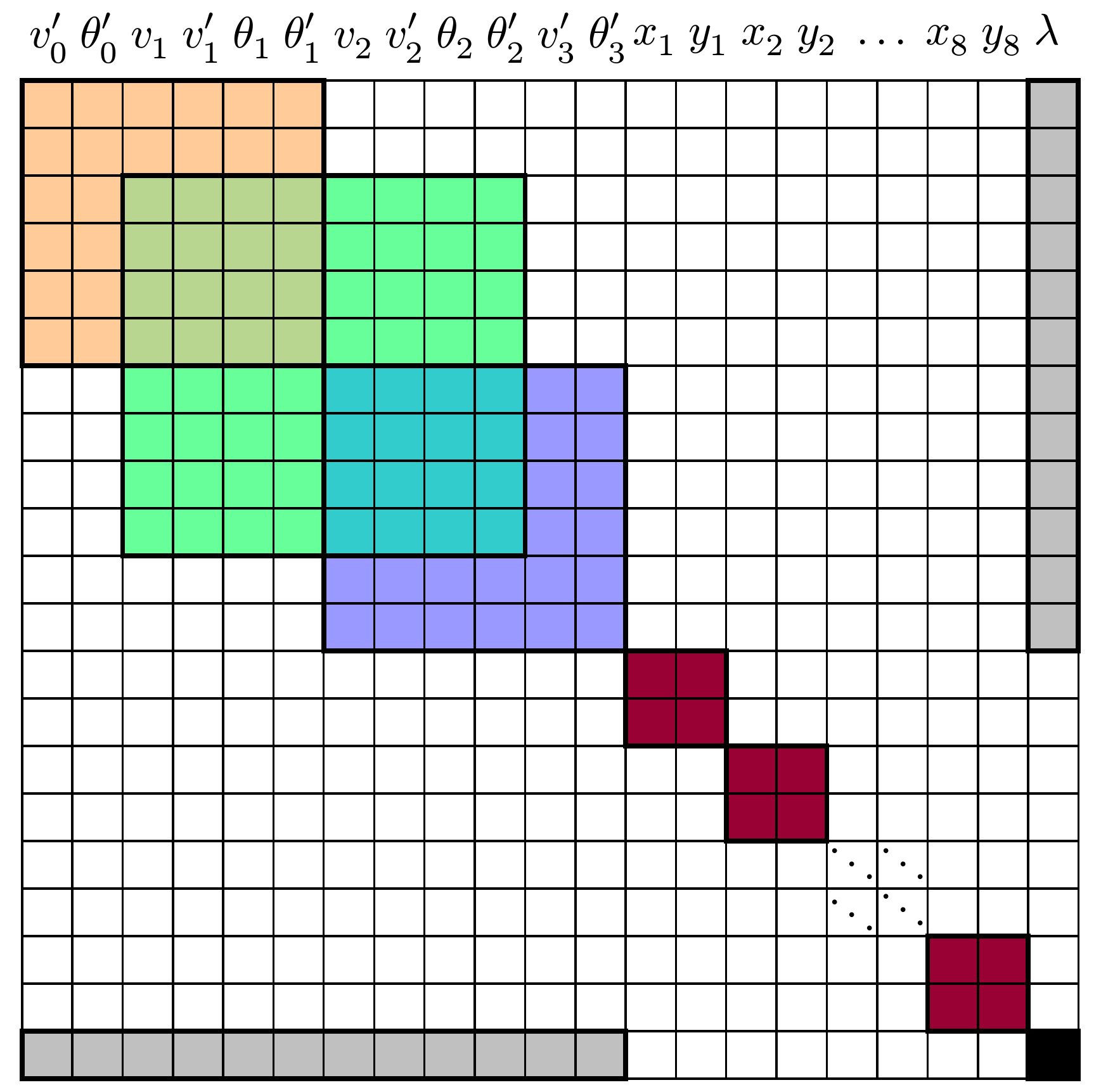}}
\end{center}
\caption
{Global connectivity structure of the finite dimensional optimization problem.} {\small (a) Some boundary
element DOFS and the first and last $\left\{x,y\right\}$ pairs, are set equal to the appropriate boundary conditions and removed from the list of unknowns ($A$). Some
boundary element DOFS are related to boundary conditions by equality constraints ($B$). Some are either related to boundary conditions via equality constraints if
speed is non-zero, or taken care of by singular elements($C$). (b) All unknowns on a node interact
with unknowns on only two neighboring nodes. Each $\left\{x,y\right\}$ pair interacts only with itself.
All DOFS on a node interact with $\lambda$.}
\label{fig:global_sample}
\end{figure}

We now describe the structure of the finite dimensional optimization problem
using a small mesh with $n = 3$ elements and $N = 10$ points for obstacle constraints
as shown in \Fig{fig:global_divisions}.  In FEM, each element provides
a small Hessian, typically dense, that relates all the DOFs present in
that element.  We have eight $v$ and $\theta$ DOFs on each element except
for the singular corner elements that have six.  \Fig{fig:global_matrix},
shows the global connectivity structure of the problem after boundary conditions are imposed
on boundary $v, \theta, x, \mbox{ and } y$ DOFs.  These are marked
$A$ in \Fig{fig:global_divisions}.
The three element matrices are added to their appropriate positions.
The DOFs marked $B$ (for $\theta'$) are constrained via equality constraints.
The DOFs marked $C$ (for $v'$) are constrained via equality constraints
if speed is non-zero.  Otherwise, it is infinite and is taken care using
singular elements.  The $x$ and $y$ DOFs do not enter the expression of
cost, hence all corresponding rows and columns are empty (zero).  The Hessian
of obstacles constraints does contain non-zero $2 \times 2$ blocks relating
$x$ and $y$ of the same point.

\subsection{Convergence on decreasing mesh size}
\label{sec:convergence}
To analyze the effect of mesh size on convergence,
we construct two examples of straight line motion, each with start and end positions as $\left\{0,0\right\}$ and
$\left\{10,0\right\}$, and start and end orientations as 0. In the first example, 
speed at both ends is 0. In the second, speed at both ends is 1. Tangential acceleration
at both ends is 0. We vary the number of elements from 2 to 128 in multiples of 2
and each time solve the discomfort minimization problem for one initial guess.
As the number of elements increases, the optimum cost found by the optimization
process decreases.  This is natural because increasing the mesh size means we're
minimizing a function in a superset of degrees of freedom.  As the number of elements
increases, the relative change in minimum cost decreases.  We compare all
costs with the cost corresponding to 128 elements. We see that the
32 elements give a cost that within 0.01\% of cost for
128 elements (when $v > 0$ on end-points).  The curve for $v = 0$
shows a lower convergence rate and we believe that the reason behind this
is using standard Gauss-Legendre quadrature for the singular elements.
A more precise procedure would use specially designed quadrature scheme
keeping in mind the form of singularity at end-points.  We have kept this as
part of future work.

\begin{figure}[t!]
\begin{center}
	\includegraphics[scale=0.3]{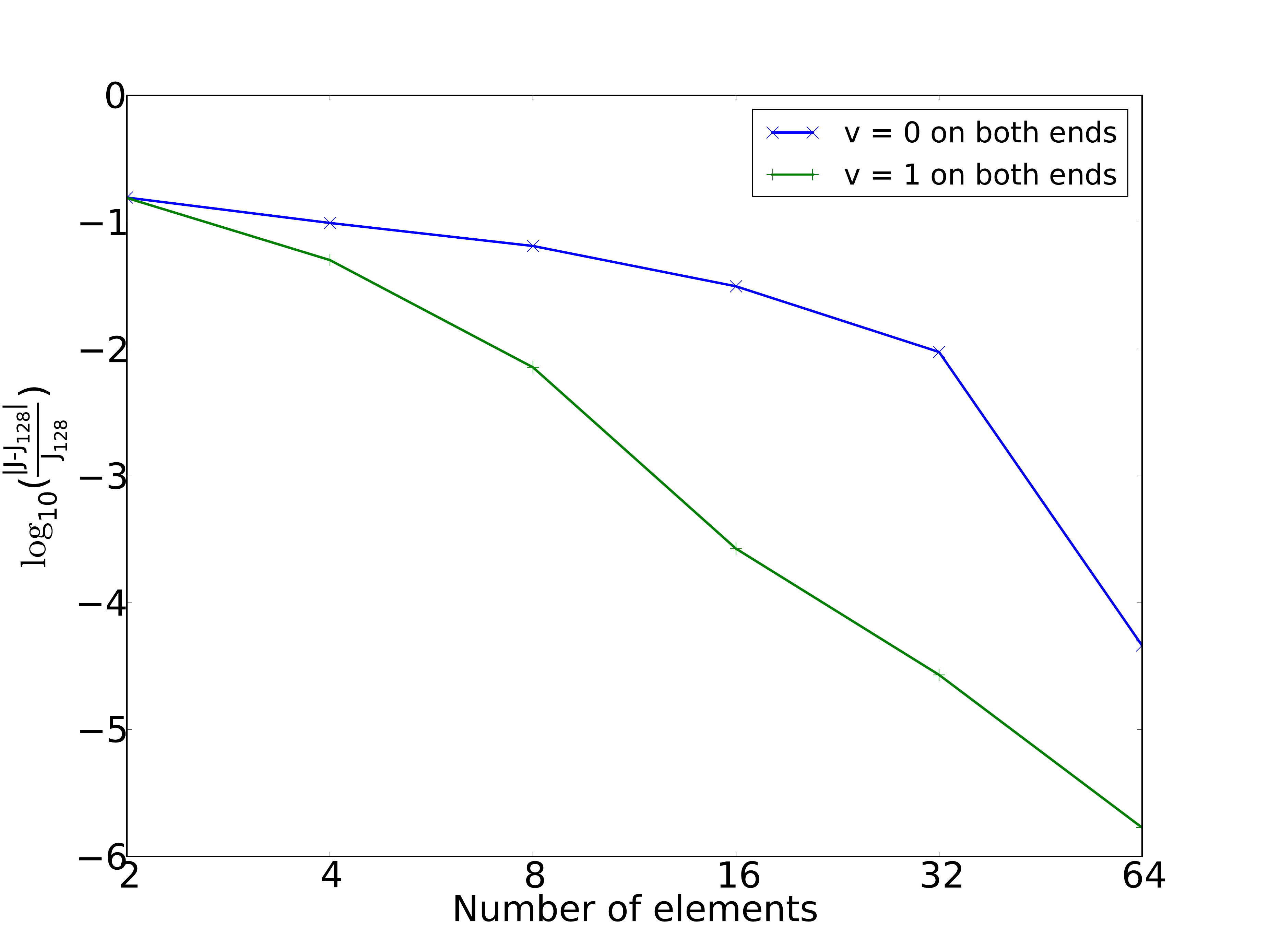}
\end{center}
\caption{Convergence on decreasing mesh size.}{\small $\log_{10}(\frac{|J - J_{128}|}{J_{128}})$ on $y-$axis against number of elements
on $\log_2$ scale on $x-$axis. The curve on top is for zero speed boundary
conditions on both ends and the curve on bottom is for unit speed boundary conditions
on both ends.}
\label{fig:convergence}
\end{figure}

\section{Initial guess for the optimization problem}

We have described a nonlinear constrained optimization
problem to find an optimal trajectory that results in a small discomfort.
Because of the non-linearity and presence of both inequality and inequality
constraints, it is crucial that a suitable initial guess of the trajectory
be computed and provided to an optimization algorithm.

Many packages can generate their own ``starting points'', but a good
initial guess that is within the feasible region can easily reduce the
computational effort (measured by number of function and derivative
evaluation steps) many times.  Not only that, reliably solving a nonlinear
constrained optimization problem without a good initial guess can be
extremely difficult.  Because of these reasons, we invest considerable
mathematical and computational effort to generate a good initial guess of the
trajectory.

\subsection{Overview}
\label{sec:iguess_overview}

As described in \Sec{sec:parameterization_of_trajectory}, a trajectory can be completely
described by its length $\lambda$, orientation $\theta(u)$, and the
speed $v(u)$ for $u \in [0,1]$.  Our optimization problem is to find the
scalar $\lambda$ and the two functions $\theta$ and $v$ that minimize the discomfort. 
We compute the initial guess of trajectory by computing $\lambda$ and $\theta$ first and
then computing $v$ by solving a separate optimization problem.  We
emphasize that the initial guess computation process must deal with
arbitrary inputs and reliably compute the initial guesses.

Before we discuss the initial guess of $\theta$, we must discuss a genuine
non-uniqueness issue. It is obvious that there exist infinitely many paths for a given
pair of initial and final orientations. There exist at least two different
kinds of non-uniqueness. The first kind of non-uniqueness exists
because multiple numerical values of an angle correspond to a single
``physical'' orientation. The second kind of non-uniqueness exists
because even for the same numerical values of initial and final angles,
one can end up in one of multiple local minima after optimization.
We now discuss these in detail.

\subsection{Multiplicity of paths}
\label{sec:iguess_overview_multiplicity}

Since the trajectory orientation $\theta$ is an angle, a single $\theta$ value
is completely equivalent in physical space to $\theta \pm 2 n \pi \; \forall n
\in \mathbb{N}$.  However, consider a trajectory that starts with a given
angle $\theta_0$, and stops at orientation $\theta_\tau$ (where $\tau$
denotes final time).  Such a trajectory will be different than a trajectory
that starts off with the same orientation but stops at $\theta_\tau \pm 2 n
\pi$.  This is because $\theta$ is continuous and cannot jump to a
different value in between.  Of course, the boundary condition will still
be satisfied.  Thus, even though the original trajectory
optimization problem is specified using a single stopping orientation, we
must consider multiple stopping orientations, differing by $2 \pi$, when
computing the initial guess as well as solving the original discomfort
minimization problem.  We have called this a ``parity'' problem.  Note that
the same logic of parity applies to the starting orientation, but what
matters is the difference and we have chosen to vary only the ending
orientation by choosing different values of $n$.

\begin{figure}[b!]
\begin{center}
\subfigure[Four paths]{\label{fig:parity_abcd_xy}
\includegraphics[scale=0.35]{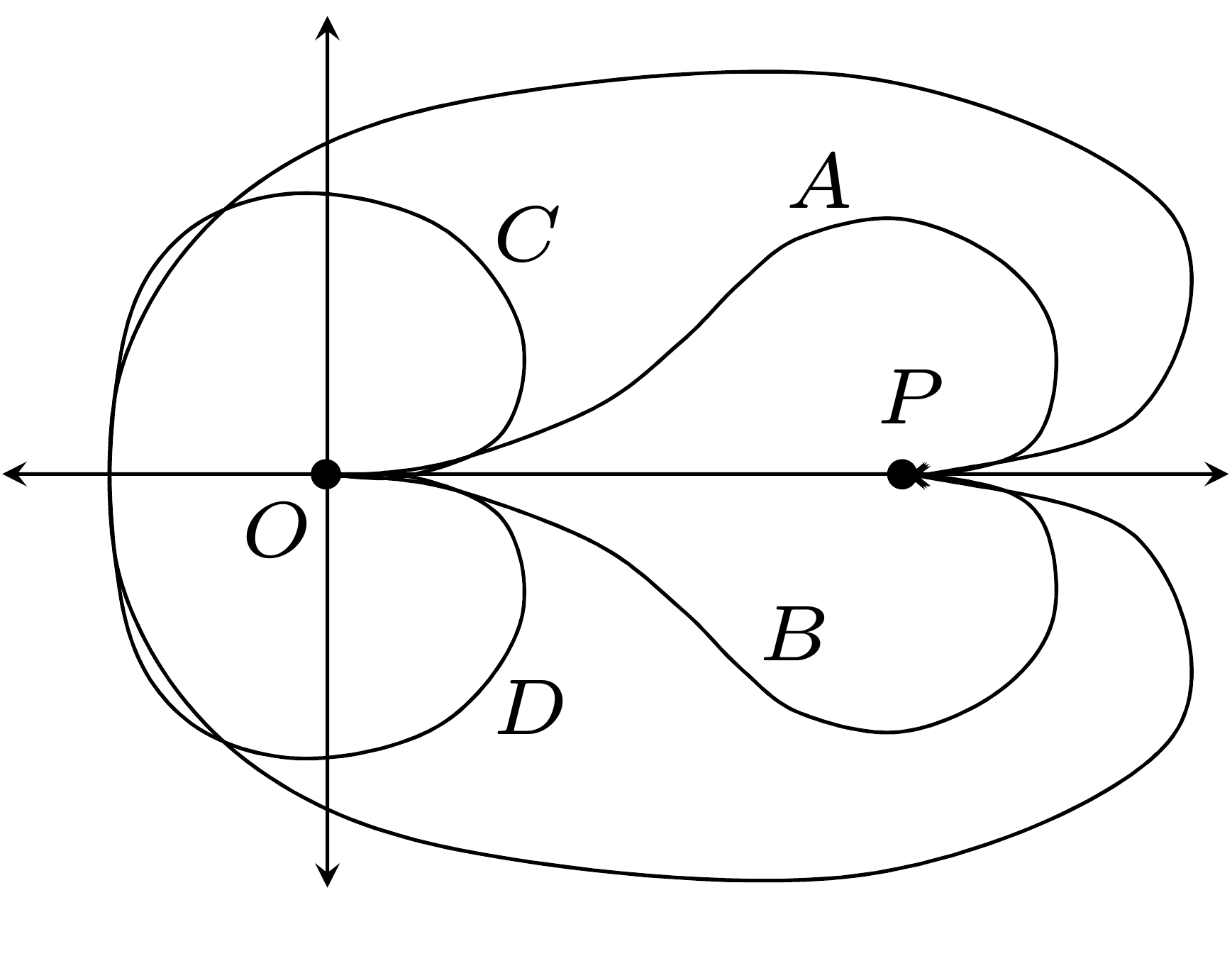}}
\subfigure[Four $\theta$ curves]{\label{fig:parity_abcd_theta}
\includegraphics[scale=0.35]{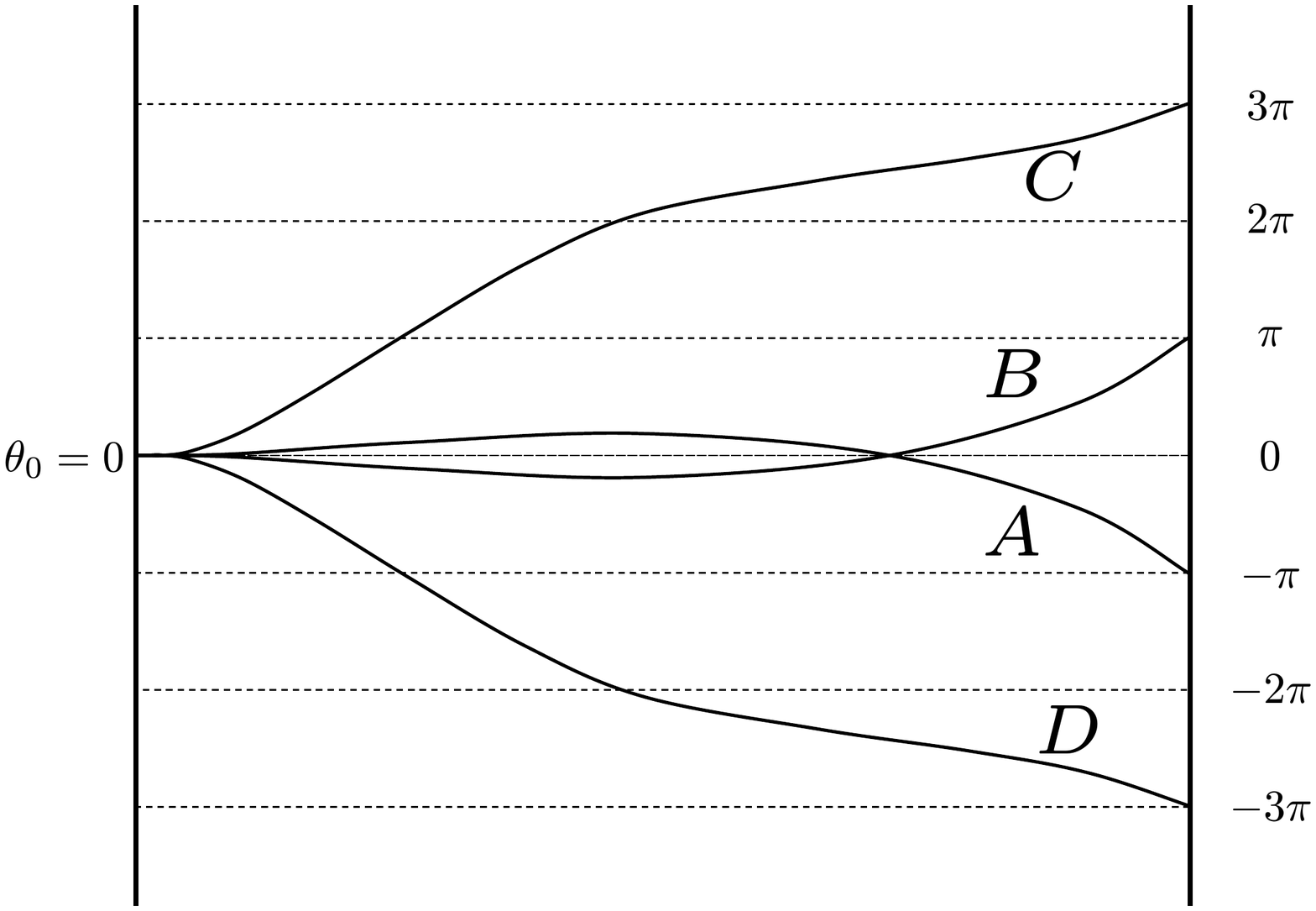}}
\end{center}
\caption{Four paths with different parity}{\small The paths $A, B, C,$ and $D$ start from
$O$ and reach $P$ at identical physical angles but, looked as $\theta$ curves,
their ending angles differ by integer multiples of $2 \pi$.}
\label{fig:parity_abcd_xy_theta}
\end{figure}

\Fig{fig:parity_abcd_xy_theta} shows a few examples of this parity.  It
shows four paths corresponding to different $n$ each sharing a common
starting angle, but reach the destination at $\{ -3\pi,-\pi,\pi,3\pi \}$.
Of course, we could create more paths by increasing the $2\pi$ difference
but doing so makes the paths more convoluted and self-intersecting in
general.  This is because when $n$ is too large in magnitude, $\theta(u)$ has
to vary rapidly at least around some $u$ values in $[0,1]$ to satisfy the
larger difference in boundary conditions.  For optimization purposes, we
assume that the starting and ending orientations are given between
$[0,2\pi)$ and we choose just three end-point orientations that give
the least difference $\left| \theta_\tau - \theta_0 \right|$.

\begin{figure}[t!]
\begin{center}
\subfigure[Two paths]{\label{fig:parity_ab_xy}          \includegraphics[scale=0.31]{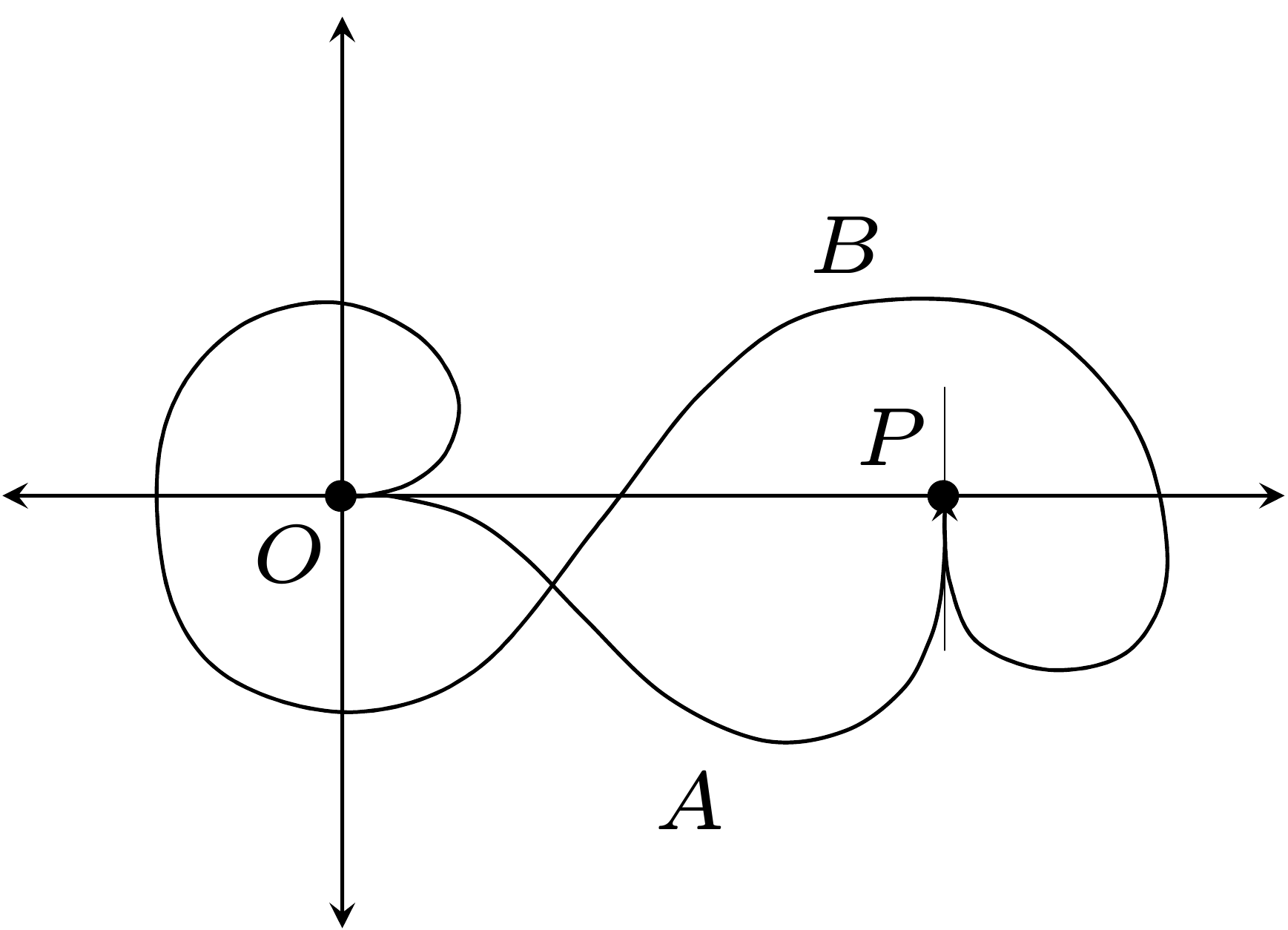}}
\subfigure[Two $\theta$ curves]{\label{fig:parity_ab_theta}       \includegraphics[scale=0.31]{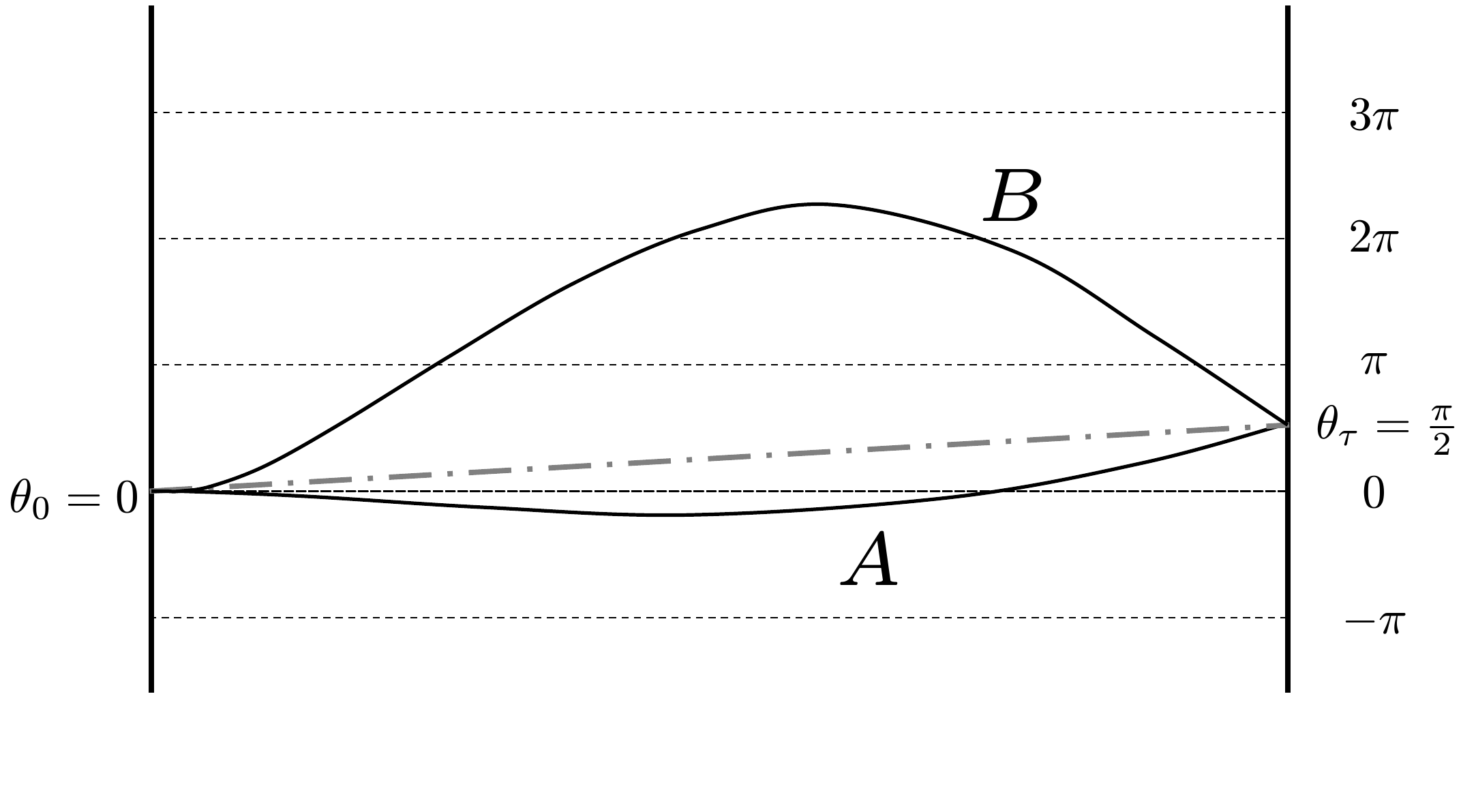}}\\
\subfigure[Deformed path]{\label{fig:parity_self_int_path}  \includegraphics[scale=0.31]{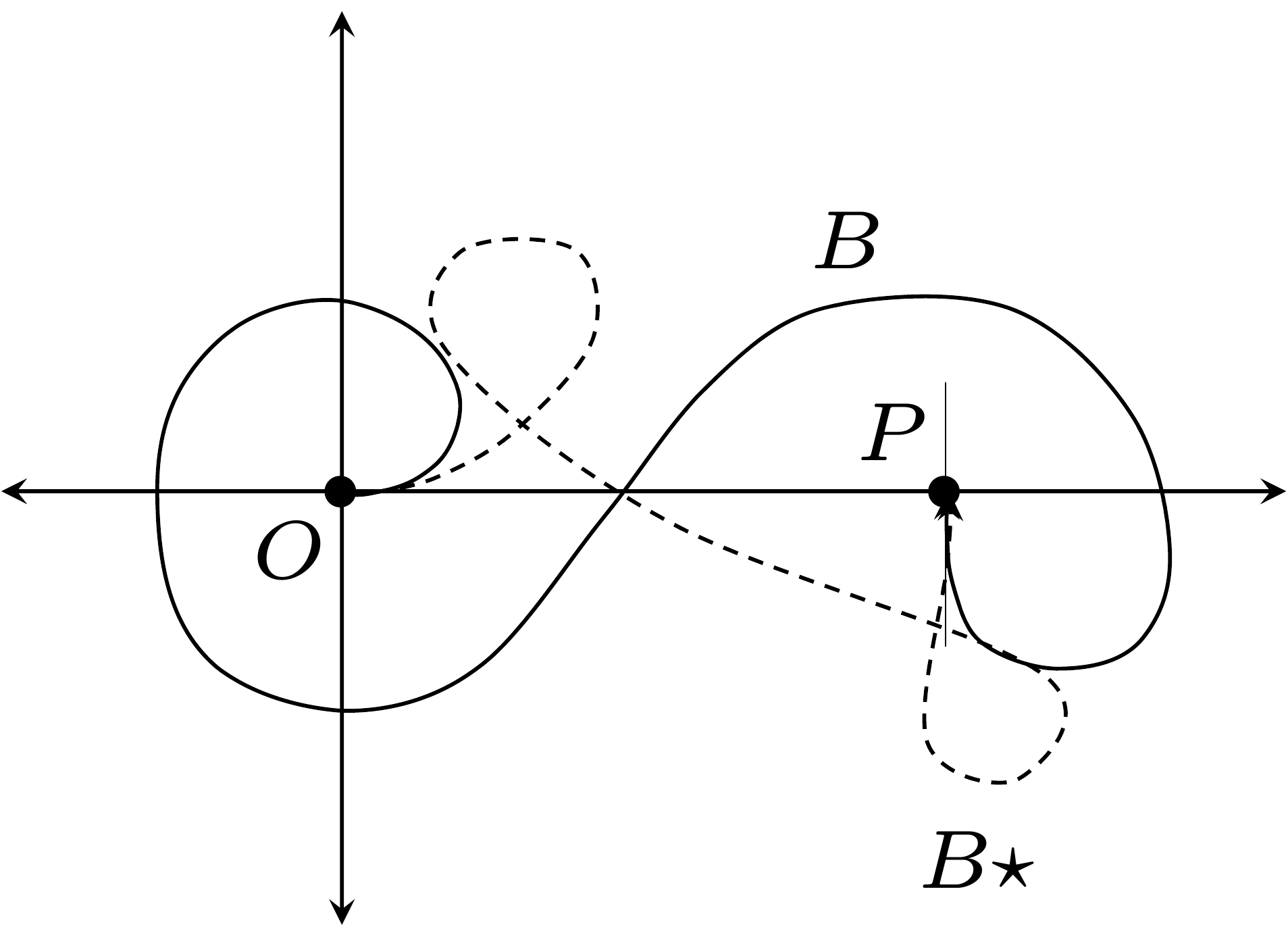}}
\subfigure[Deformed $\theta$ curve]{\label{fig:parity_self_int_theta} \includegraphics[scale=0.31]{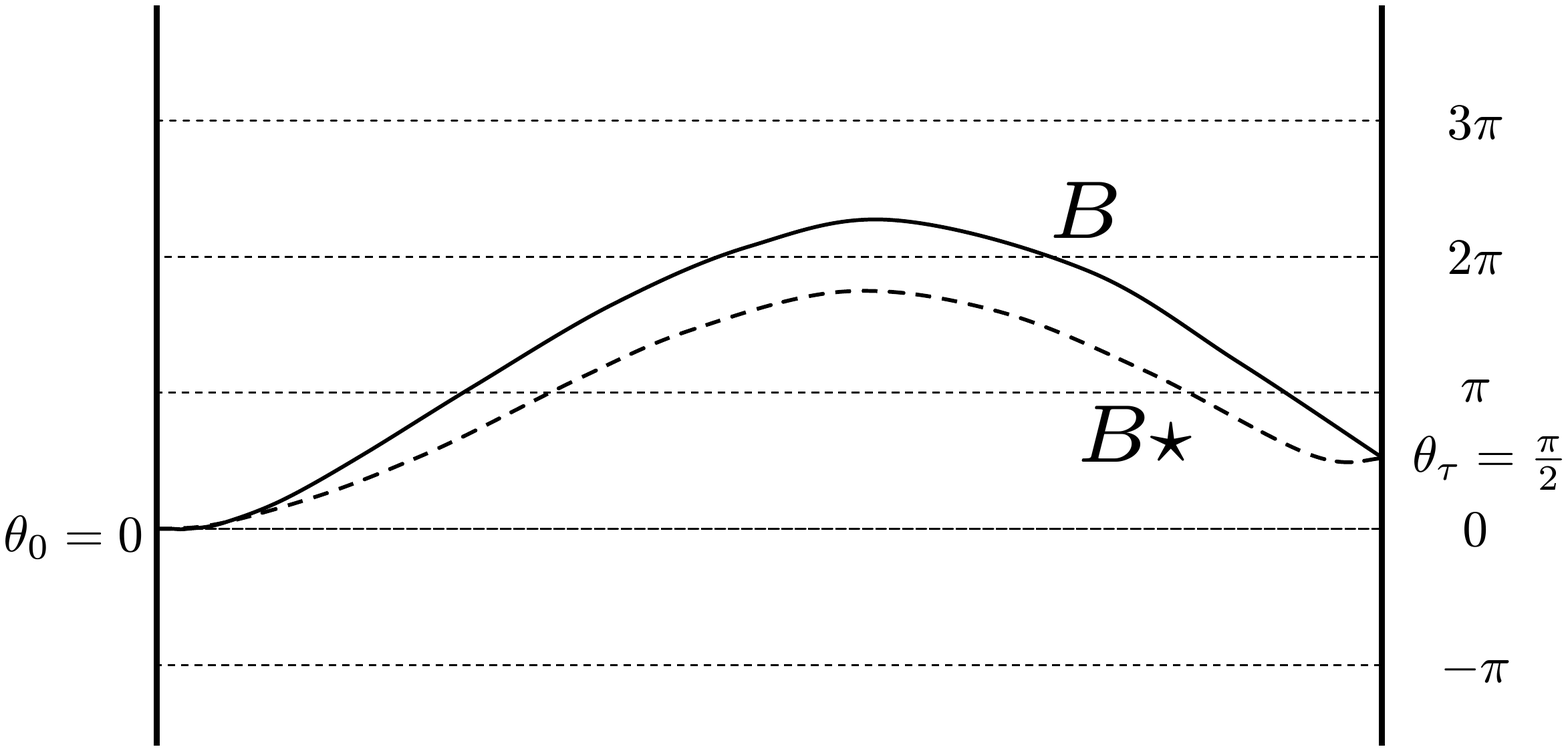}}
\end{center}
\caption{(Two local minima for same boundary conditions}{\small (a)(b) The paths
$A$ and $B$ are different but both minimize discomfort
compared to neighboring paths.  (c)(d) $B_*$ is obtained by a continuous
deformation of $B$. The $\theta$ curves of $B$ and
$B*$ are similar. The corresponding paths show that $B*$ contains two
self intersections and is topologically different from $A$ that does
not contain self intersections.}
\label{fig:parity_self_int}
\end{figure}

Apart from the multiplicity of $\theta$ curves due to parity, the
optimization problem discussed \Sec{sec:iguess_path} to compute $\theta$ as
well as the full discomfort minimization problem can lead to multiple
solutions even when parity remains unchanged.  This occurs due to the
non-linearity of constraints in~\Eq{eq:r_by_integral_constr}.
\Fig{fig:parity_self_int}(a) and (b) shows two such paths $A$ and $B$ that start
and end at the same numerical orientation but are qualitatively different.
We do observe such multiple minima in practice.  If we observe
$B$ more carefully, it is seen that it can be continuously
deformed into $B*$ shown in~\Fig{fig:parity_self_int}(c) and (d).
This figure clearly shows that the reason $B$ is qualitatively
different from $A$ is because of two self-intersections $-$ first
in anti-clockwise direction and second in clockwise direction.
Both ``loops'' cancel each others' changes in orientation.  We suspect
that this topological difference is the cause of multiple local minima.

Of course, this argument can be carried further and one can
introduce an equal number of clockwise and anti-clockwise loops
in arbitrary order and the final orientation will remain unchanged.
Thus, we believe that there can be infinitely many local minima.
Obviously, doing so would increase the discomfort in general and
such a path will not be desirable.  We try to avoid this problem
by setting bounds on maximum and minimum $\theta$ when we compute
the initial guess of $\theta$.  However, it is important
to not ignore multiple minima if they are found within these bounds.
If obstacles are present so that $A$ is infeasible, $B$ might be
chosen even though it is longer and has more turns.

Thus, because of these two kinds of multiplicities, we use more than one
initial guess when minimizing the discomfort and choose the one that has
the minimum discomfort and satisfies the constraints.  We discuss
the details in the following section.

\subsection{Initial guess of path}
\label{sec:iguess_path}

We compute initial guesses for $\lambda$ and $\theta(u)$ using two
different methods.  The first method computes a $\theta(u)$ such that the
trajectory has a piecewise constant curvature.  This is a
computationally inexpensive method and does not satisfy many of the
constraints exactly.  The output of this method can be used to solve the
full discomfort minimization problem.

The second method computes a $\theta(u)$ and $\lambda$ by solving an
auxiliary (but simpler) nonlinear constrained optimization problem.  Of
course, now we need an initial guess for this new optimization problem!
The output of the ``constant curvature'' method mentioned above is used as
the initial guess.  Unlike the first method, the output of this second
method leads to trajectories that have continuous and differentiable curvature 
and also
satisfy boundary conditions and maximum curvature constraint
exactly.

\subsubsection{Piecewise constant curvature path}
\label{sec:iguess_path_cc}

In the full discomfort minimization problem, the orientation $\theta(u)$ has to
satisfy the boundary conditions and \Eq{eq:r_by_integral_constr}. In total, there are four constraints $-$ two
linear (those due to boundary conditions) and 
two non-linear (those of \Eq{eq:r_by_integral_constr}).  

For computing initial guess of $\theta$, 
we modify the inputs of the full optimization problem using a rotation such that
initial and final position have the same $y$ coordinate value.  The
initial and final orientations are also modified appropriately.  Once we
find an initial guess for the transformed input, it can be easily
transformed back to the original configuration by the inverse rotation. 
This is done to allow efficient storage of precomputed $\theta$ guesses
for various end-point conditions.

Thus, the inputs to the initial guess generation problem  are the
initial and final positions, $x_0$ and $x_\tau$, and orientations, 
$\theta_0$ and $\theta_\tau$, in the rotated frame.
The output will be a path length $\lambda$ and a function $\theta(u)$. 

We begin by choosing the value of path length $\lambda$ as max$(R, 2*\Delta L)$,
where $R$ is the minimum turning radius of the robot and 
$\Delta L = \norm{\ve{r}_\tau - \ve{r}_0}$.  Using this maximum takes
care of the case when initial and final positions are very close
to each other. In such a case, the path length is decided by the minimum turning
radius constraint.

\begin{figure}
\begin{center}
\subfigure[Cost from~\Eq{eq:init_guess_cc}]{\label{fig:constr_cost}          \includegraphics[scale=0.45]{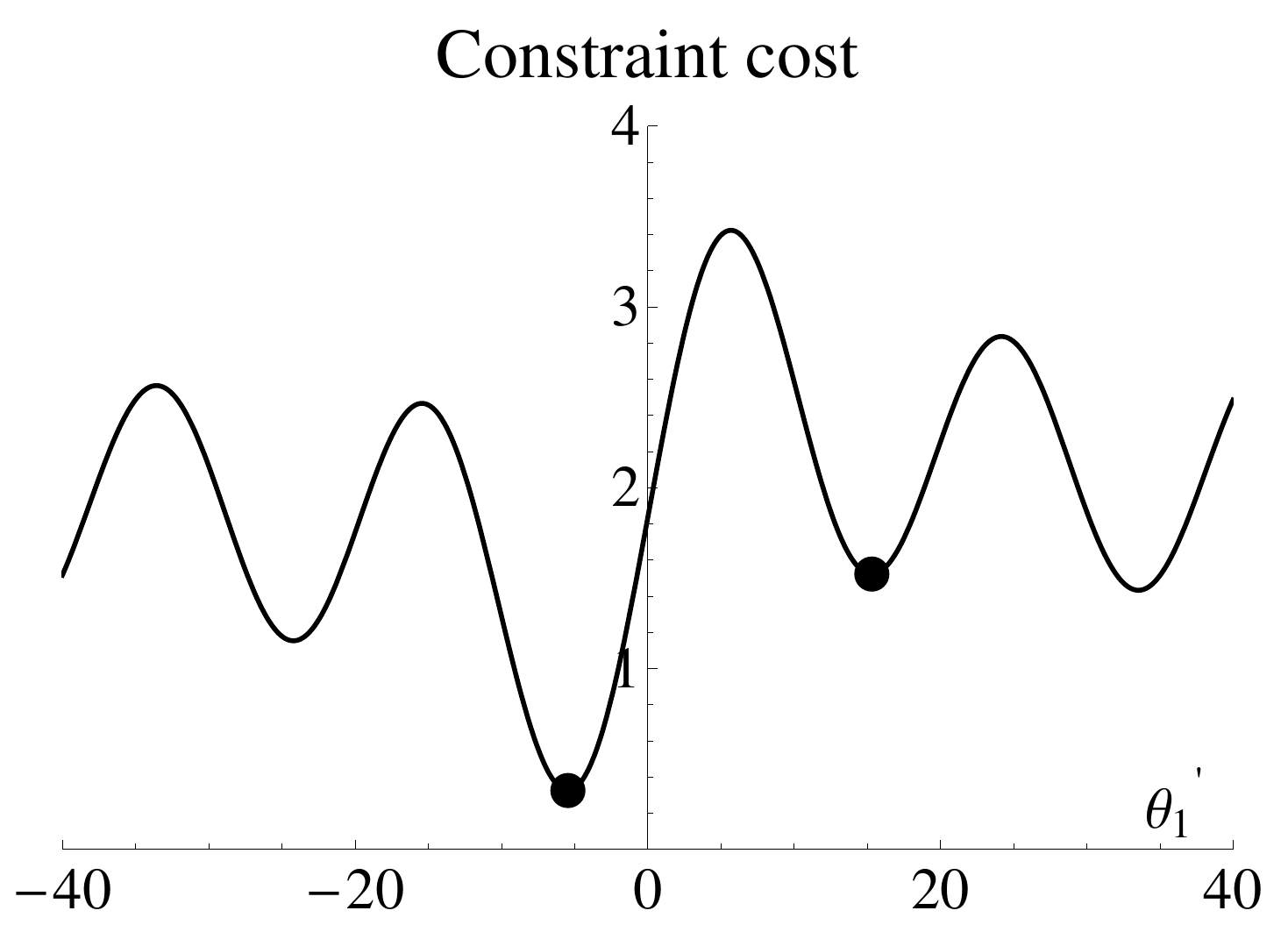}}
\subfigure[Two constant curvature optimizers]{\label{fig:two_theta_guesses}    \includegraphics[scale=0.45]{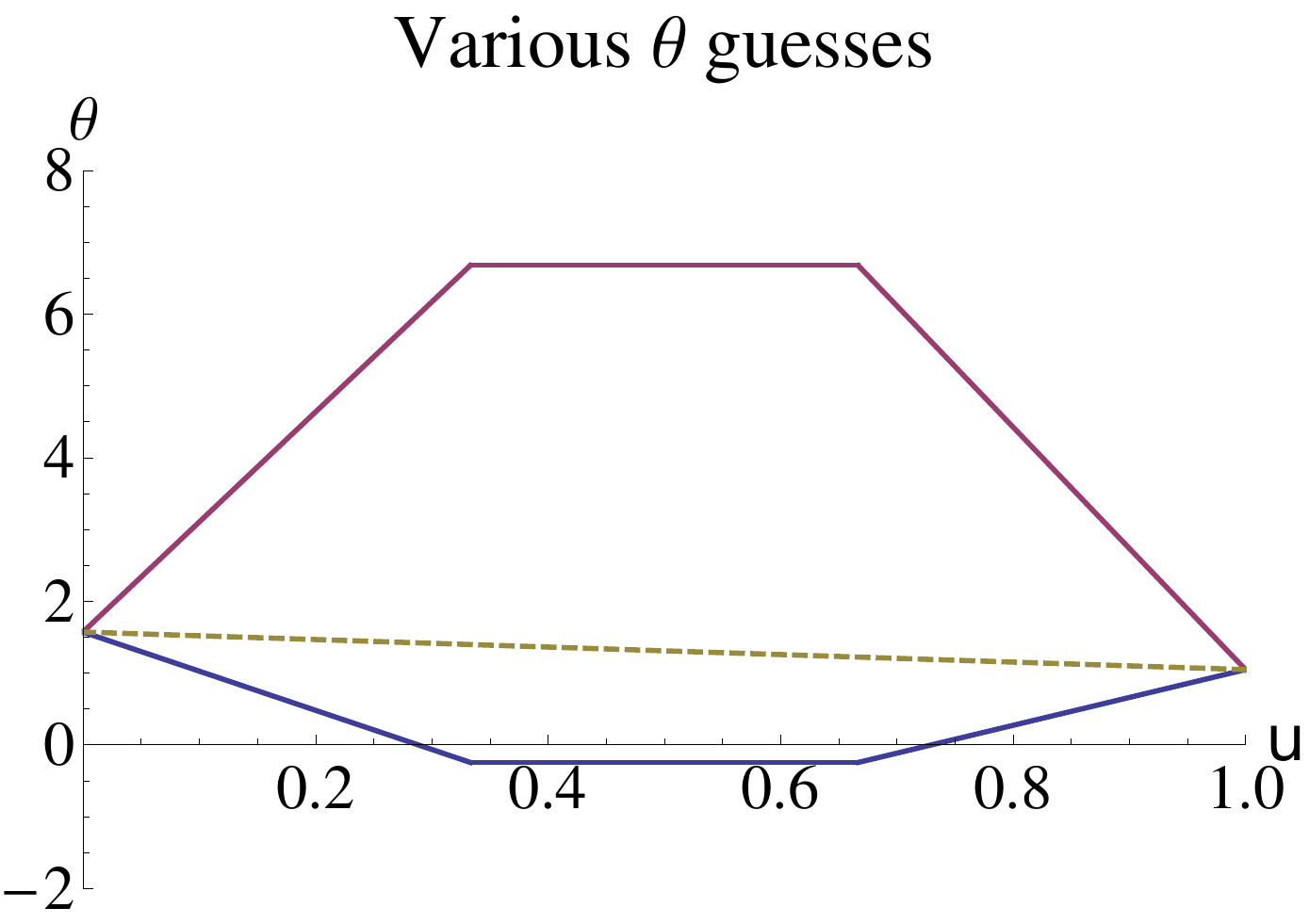}}\\
\subfigure[Paths corresponding to (b)]{\label{fig:two_paths}            \includegraphics[scale=0.4]{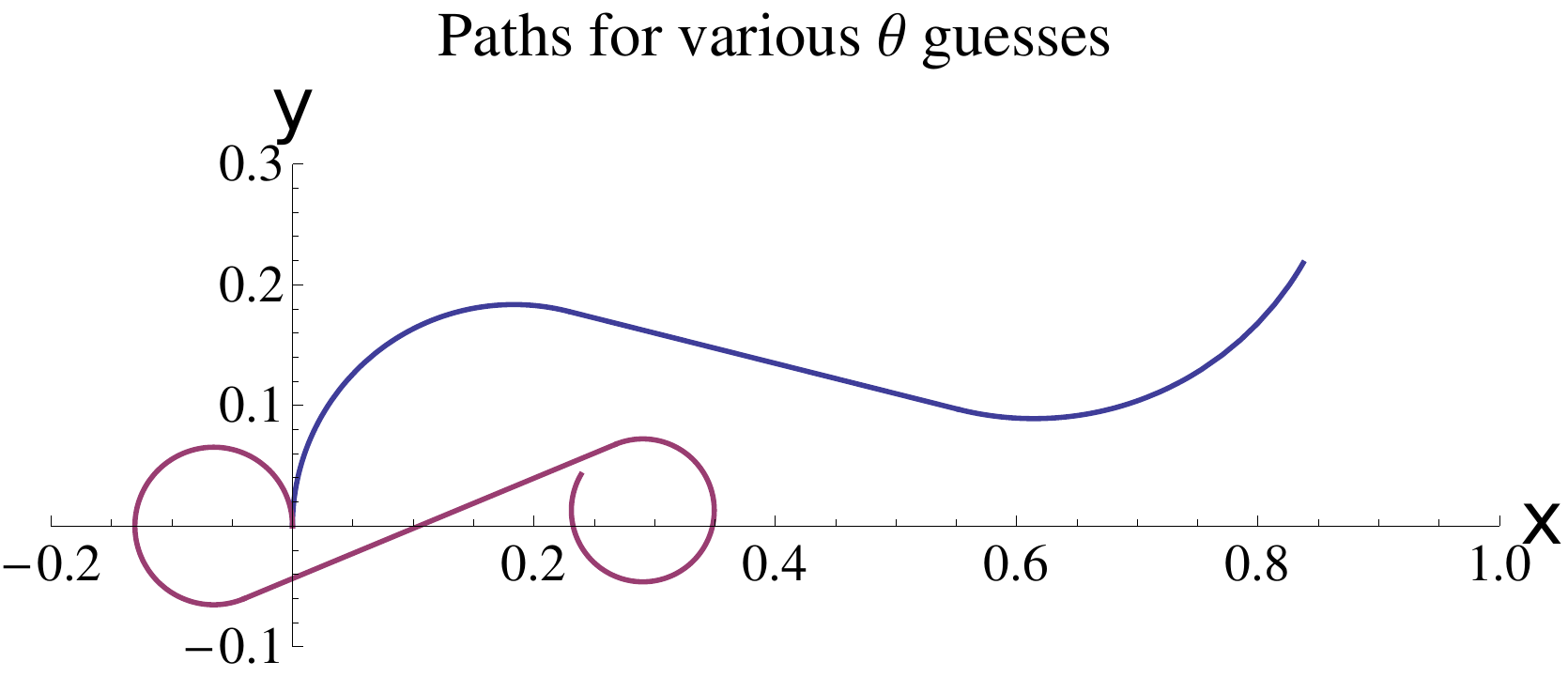}}
\subfigure[Multiple local minima]{\label{fig:colored_theta_curves} \includegraphics[scale=0.4]{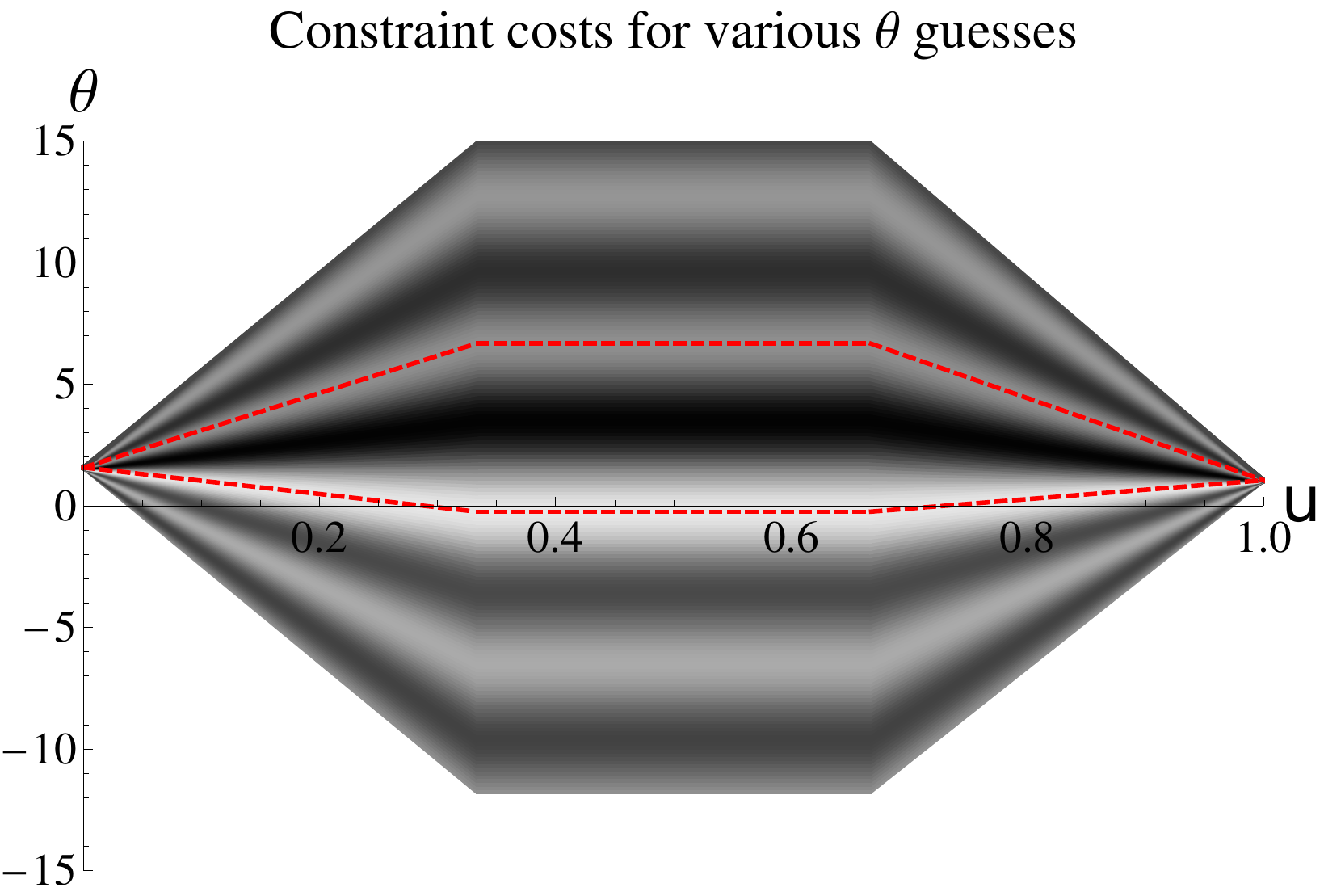}}
\end{center}
\caption{Piecewise constant curvature initial guesses}{\small (a) Multiple local minima in graph of
$J_{cc}$ of \Eq{eq:init_guess_cc}. Two minima closest to the maxima are highlighted.
(b) Piecewise constant curvature paths (not dashed) corresponding to highlighted minima in (a).
(c) Paths corresponding to the $\theta$ curves in (b).  Both start at $\frac{\pi}{2}$
and end at $\frac{\pi}{3}$. (d) Lighter shade represents minima and darker shade represents
maxima.  The two optimizing paths of (b) are shown.}
\label{fig:constr_cost_all}
\end{figure}

Ideally, an initial guess of $\theta(u)$ should obey the following constraints
so that the constraints of the full optimization problem are satisfied:
%--------------
\begin{equation}
\label{eq:theta_linear_constr}
\begin{split}
\theta(0) &= \theta_0, \\
\theta(1) &= \theta_{\tau},
\end{split}
\end{equation}
%--------------
and transformed \Eq{eq:r_by_integral_constr}
%--------------
\begin{equation}
\label{eq:theta_nonlinear_constr}
\begin{split}
\lambda \int_0^1 \cth\, d u &= x_{\tau} - x_0,\\
\lambda \int_0^1 \sth\, d u &= 0,
\end{split}
\end{equation}
%--------------
Consider a piecewise linear function that looks like the solid curves
in~\Fig{fig:constr_cost_all}(b).

Essentially, each such curve is defined on
$[0,1]$, is continuous, and is made of three line segments in $[0,\frac{1}{3}]$,
$[\frac{1}{3},\frac{2}{3}]$, and $[\frac{2}{3},1]$.  The middle line segment
has zero derivative.
The values at 0 and 1 are known and the only variable is the function value
on the middle segment. Equivalently, one can use the slope of first line
segment as the variable. Let this slope be denoted by $\theta'_1$. Then, 
we define $\theta(u)$ as
%--------------
\begin{equation}
 \theta(u) = \left\{ \begin{array}{rll}
\theta_0 + \theta'_1 u  & \mbox{if} & 0 \leq u < \frac{1}{3}; \\
\theta_0 + \frac{1}{3} \theta'_1  & \mbox{if} & \frac{1}{3} \leq  u < \frac{2}{3}; \\
 \theta_0 + \frac{1}{3} \theta'_1 - 3(\theta_0 - \theta_1 + \frac{1}{3}\theta'_1) (u - \frac{2}{3}) & \mbox{if} &  \frac{2}{3} \leq  u \leq 1.
\end{array}\right.
\end{equation} 
%--------------
If we use such a curve for $\theta(u)$, it will
result in a circular arc, a tangent line segment, and another circular arc
tangential to the middle segment, in that order.  This, in turn, implies
that the resulting path will have a piecewise constant curvature.

To determine $\theta(u)$, we need to determine the value of the unknown
slope $\theta'_1$. Since only one value cannot satisfy two
constraints of \Eq{eq:theta_nonlinear_constr}, we minimize
\begin{equation}
\label{eq:init_guess_cc}
J_{cc}(\theta'_1) = \left(\int_0^1 \cth\, d u - 1 \right)^2 + \left(\int_0^1 \sth\, d u  \right)^2
\end{equation}
to find $\theta'_1$. \Fig{fig:constr_cost_all}(a) shows
the plot of $J_{cc}$ as a function of $\theta'_1$.  Depending on
the boundary conditions, the shape of $J_{cc}$ changes but qualitatively
it has the behavior as shown $-$ oscillatory with a maximum not too
far from zero.  We find this maximum using a table lookup and
the neighboring two minima to compute the initial guess.
\Fig{fig:constr_cost_all}(c) shows two paths using this method
where $\theta_0 = \frac{\pi}{2}$ and $\theta_\tau = \frac{\pi}{3}$.
The path length is 1.  As seen, the curve end-point is not too
far from $x-$axis, and the curve satisfies the boundary condition
on $\theta$. \Fig{fig:constr_cost_all}(d) shows
the cost $J_{cc}$ for various constant curvature paths.  The lighter
shade corresponds to the minima.

\subsubsection{Optimization approach for initial guess of path}
\label{sec:iguess_path_opt}

In this second method to compute the initial guess of the path,
we minimize
\begin{equation}
\label{eq:th_init_guess_opt}
J(\theta, \lambda) = \lambda + w \int_0^1 \theta''^2 d u
\end{equation}
where $w := \max(\Delta L, R)$, and $\theta$ must satisfy the
boundary conditions, the two equality constraints of
\Eq{eq:r_by_integral_constr}, and the curvature constraint 
$$
\left| \theta'(u) \right| \le \lambda \kappa_{\max} \; \forall u \in [0,1].
$$
We do not impose the obstacle related constraints in this problem.
This problem is related to the concept of ``Minimum Variation Curves''
\cite{Moreton:CSD_93_732} which have been proposed for curve shape design.
We add the curve length $\lambda$ so that in the presence of
multiplicities, discussed in \Sec{sec:iguess_overview_multiplicity} earlier,
the curves with smaller lengths are preferred.
This optimization problem is discretized using $C^1$ finite elements as
described in \Sec{sec:element_shape_functions}, and the initial guess is the piecewise
constant curvature function from \Sec{sec:iguess_path_cc}.

%\clearpage
\begin{figure}[hb!]
\begin{center}
	\includegraphics[scale=0.45]{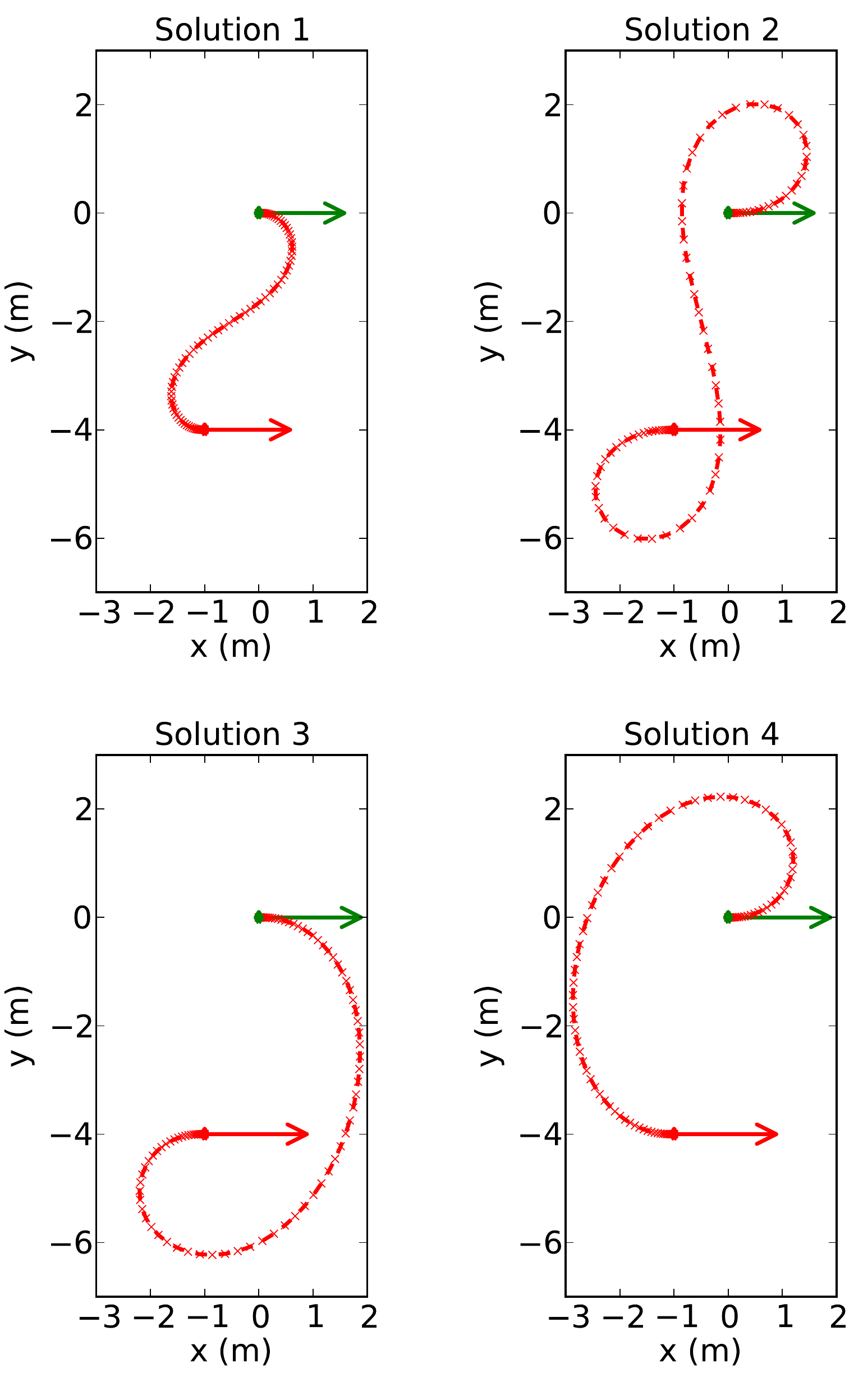}
\end{center}
\caption{Four initial guess of path.}{\small{Problem input is as follows: initial position = $\left\{0,0\right\}$, orientation = 0,
speed = 0, tangential acceleration = 0; final position = $\left\{-1,-4\right\}$, orientation = 0,
speed = 0, tangential acceleration = 0. The four initial guesses of path are computed
using the method described in \Sec{sec:iguess_path} so that final orientation 
in (a),(b),(c) and (d)
is 0, 0, $-2\pi$ and $2\pi$ respectively. All quantities have appropriate units in terms 
of meters and seconds. Initial and final positions are shown by markers and
orientations are indicated by arrows. While the path is parameterized by $u$, for
ease of visualization, we show markers at equal intervals of time. Thus distance between
markers is inversely proportional to speed. }}
\label{fig:path_init-s_shape}
\end{figure}

\subsection{Initial guess of speed}
\label{sec:iguess_speed}

Computing the initial guess of $v$ is relatively simpler.  We solve a
convex quadratic optimization problem with linear inequality box
constraints to compute the initial guess.  Because of convexity
of the functional and the convex shape of the feasible region, this
problem has a unique solution and an initial guess is not necessary
to solve it. Any good quality optimization package can find the solution without
an initial guess.  Of course, because of the simplicity of box constraints, we
can and do provide a feasible initial guess for this auxiliary problem.

First consider the case when both end-points have non-zero speed.
We minimize
\begin{equation}
\label{eq:v_init_guess}
J(v) = \int_0^1 v''^2 d u
\end{equation}
subject to boundary constraints $v(0) = v_0 > 0$, $v(1) = v_1 > 0$,
$v'(0) = \frac{a_0 \lambda}{v_0}$, $v'(1) = \frac{a_1 \lambda}{v_1}$
and inequality constraints $v_{\min}(u) \le v(u) \le v_{\max}(u)$
and $A_{\min}(u) \le v'(u) \le A_{\max}(u)$.
The expressions for $v'(0)$ and $v'(1)$ come from the relation
in~\Eq{eq:aT}. The length $\lambda$ is computed when the initial
guess for $\theta$ is computed.  Here we choose $v_{\min}(u) = \min(v_0, v_1)/2$
and $v_{\max}(u)$ is a constant that comes from
the hardware limits.  The function $A_{\min}(u)$ is chosen to
be the constant $10 a_{\min} \lambda / \min(v_0, v_1)$ where
$a_{\min}$ is the minimum allowed physical acceleration.
$A_{\max}(u)$ is chosen similarly using $a_{\max}$.

This optimization problem is discretized
using $C^1$ finite elements as described in \Sec{sec:element_shape_functions}
and leads to a convex programming problem that is easily solved.
Of course, this method does not take care of cases in which one or
both points have zero speed boundary conditions.

If both end-points have zero speeds, the function
\begin{equation}
\label{eq:zero_v_init_guess_function}
v(u) = v_{\max} \left( 4 u (1 - u) \right)^{2/3}
\end{equation}
satisfies the boundary conditions and singularities and has a maximum
value of $v_{\max}$.  This case does not require any optimization.

If only one of the end-points has a zero speed boundary condition,
we split the initial guess for $v$ into a sum of two functions.
The first one takes care of the singularity and the second takes
care of the non-zero speed boundary condition on the other end-point.
We now maintain only the $v_{\max}$ constraint because
$v'(u)$ is unbounded and $v_{\min} = 0$ naturally.
If the right end-point has zero speed, we choose
$$
v(u) = v_{\mbox{singular}}(u) + v_{\mbox{non-singular}}(u)
$$
where
$$
v_{\mbox{singular}}(u) = \frac{16}{9} 2^{1/3} v_{\max} u^2 (1-u)^{2/3}.
$$
This function has the correct singularity behavior and its maximum
value is $v_{\max}/2$.  The non-singular part is computed via optimization
so that the sum is always less than $v_{\max}$.  For the other case,
when left end-point has zero speed, the singular part (using symmetry) is
$$
\frac{16}{9} 2^{1/3} v_{\max} (1-u)^2 u^{2/3}.
$$

\begin{figure}[p]
\begin{center}
\subfigure[$v_{\min} = 0.5$ is active]{\label{fig:v_min_no_sing}    \includegraphics[scale=0.25]{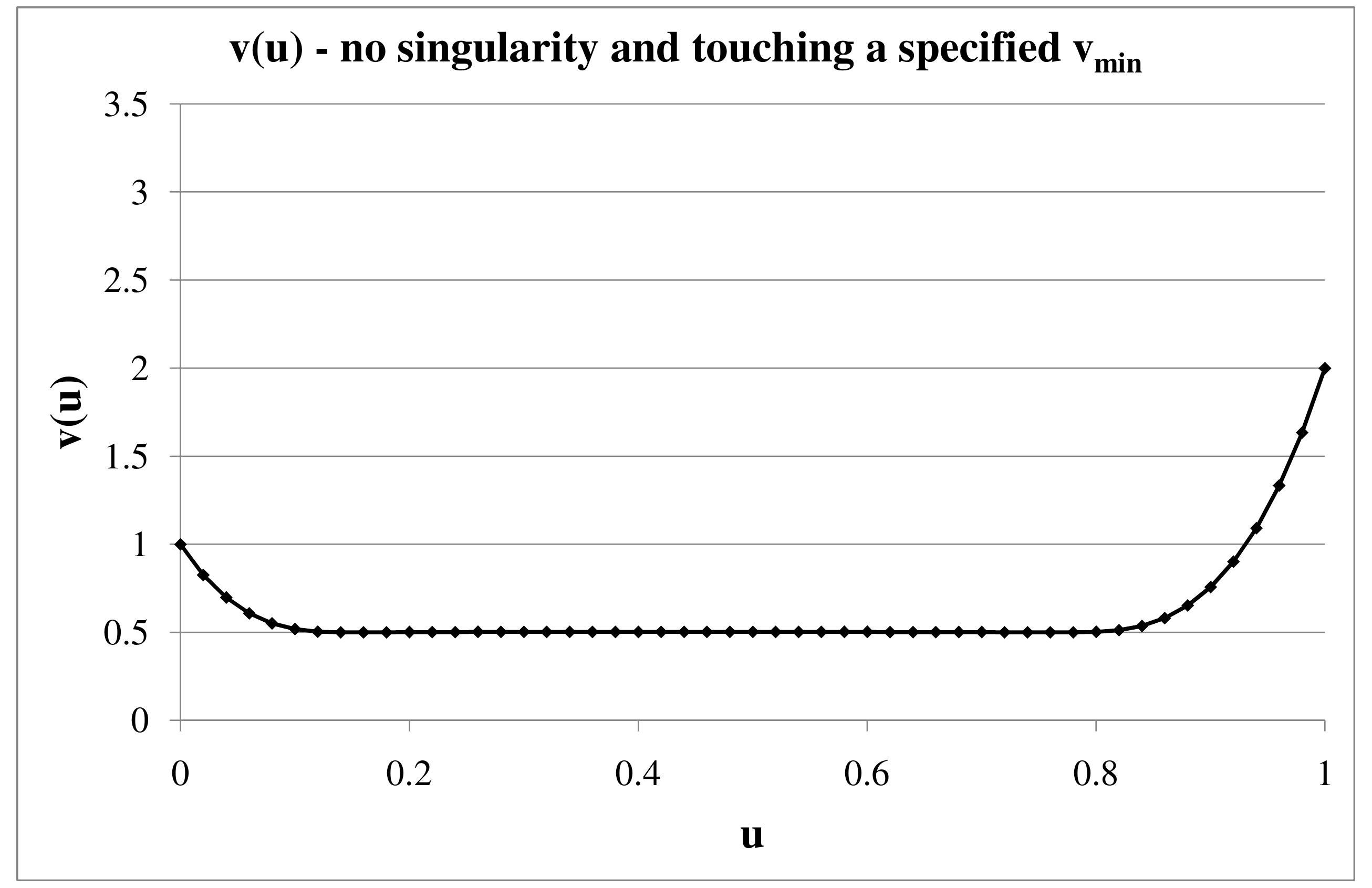}}
\subfigure[$v_{\max} = 3$ is active]{\label{fig:v_max_no_sing}    \includegraphics[scale=0.25]{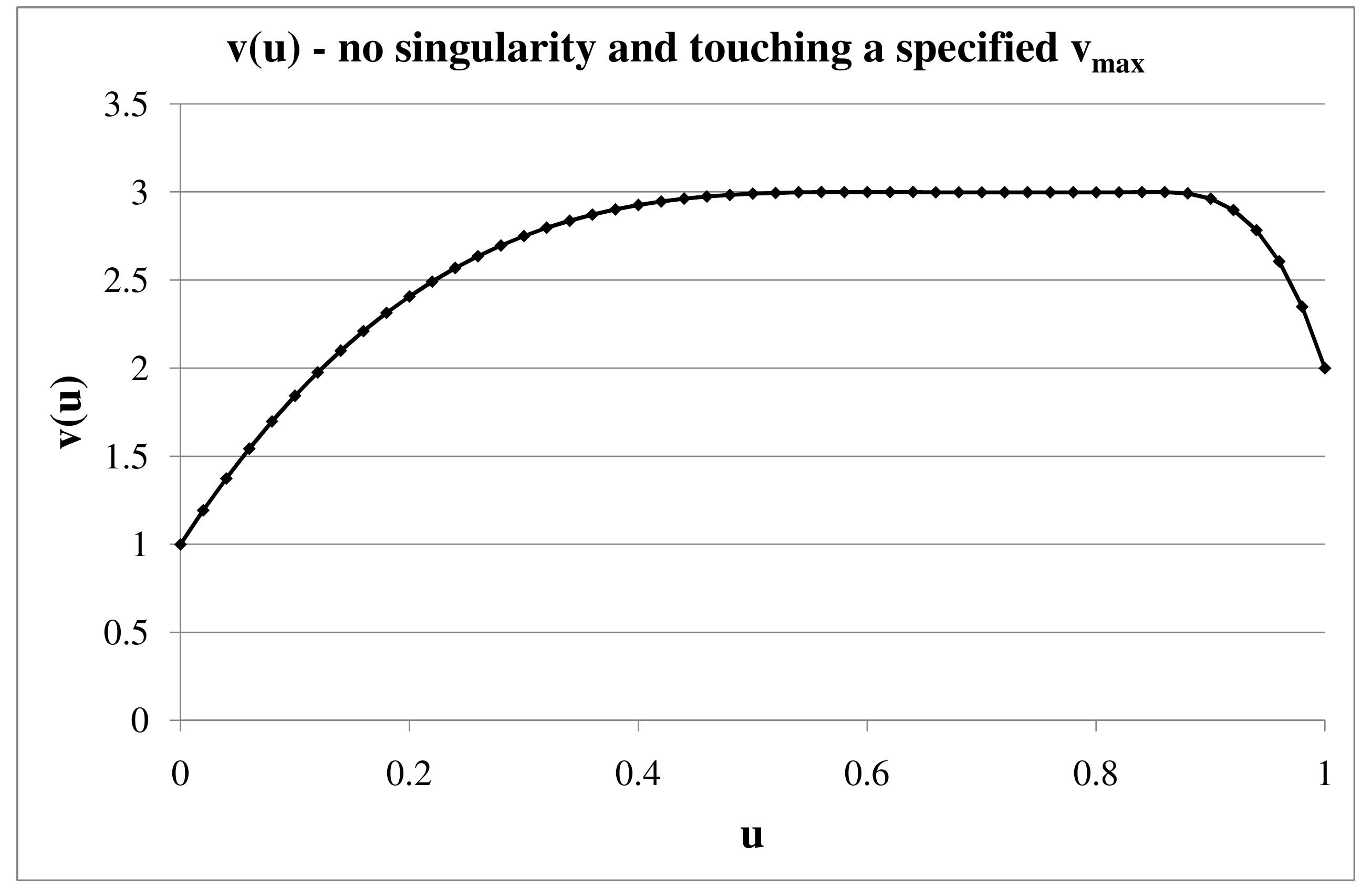}}\\
\subfigure[Both $v_{\min}$ and $v_{\max}$ active]{\label{fig:both_no_sing}     \includegraphics[scale=0.25]{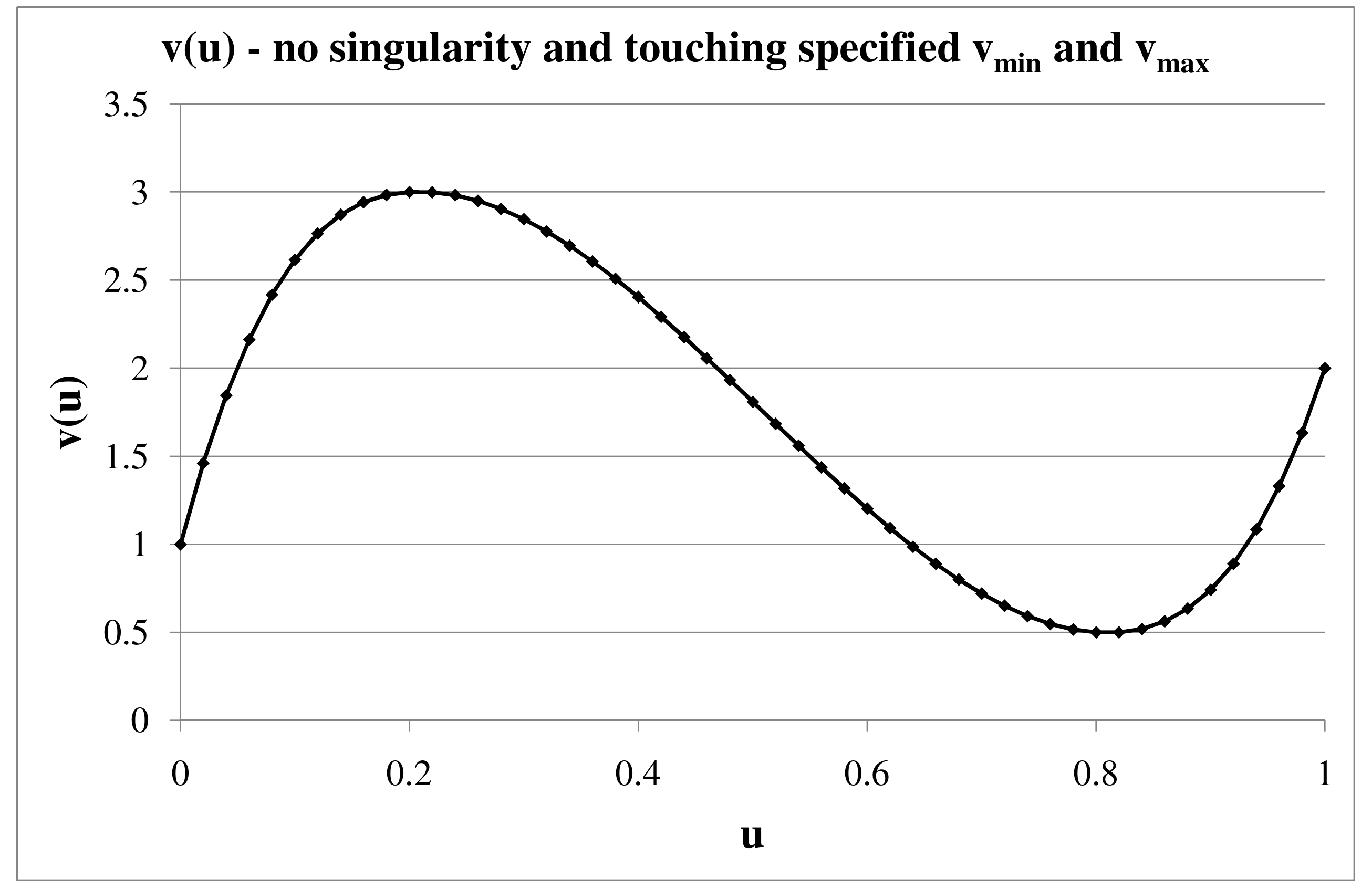}}
\subfigure[Both ends singular and $v(u) = v_{\max} \left( 4 u (1 - u) \right)^{2/3}$]{\label{fig:both_sing}        \includegraphics[scale=0.25]{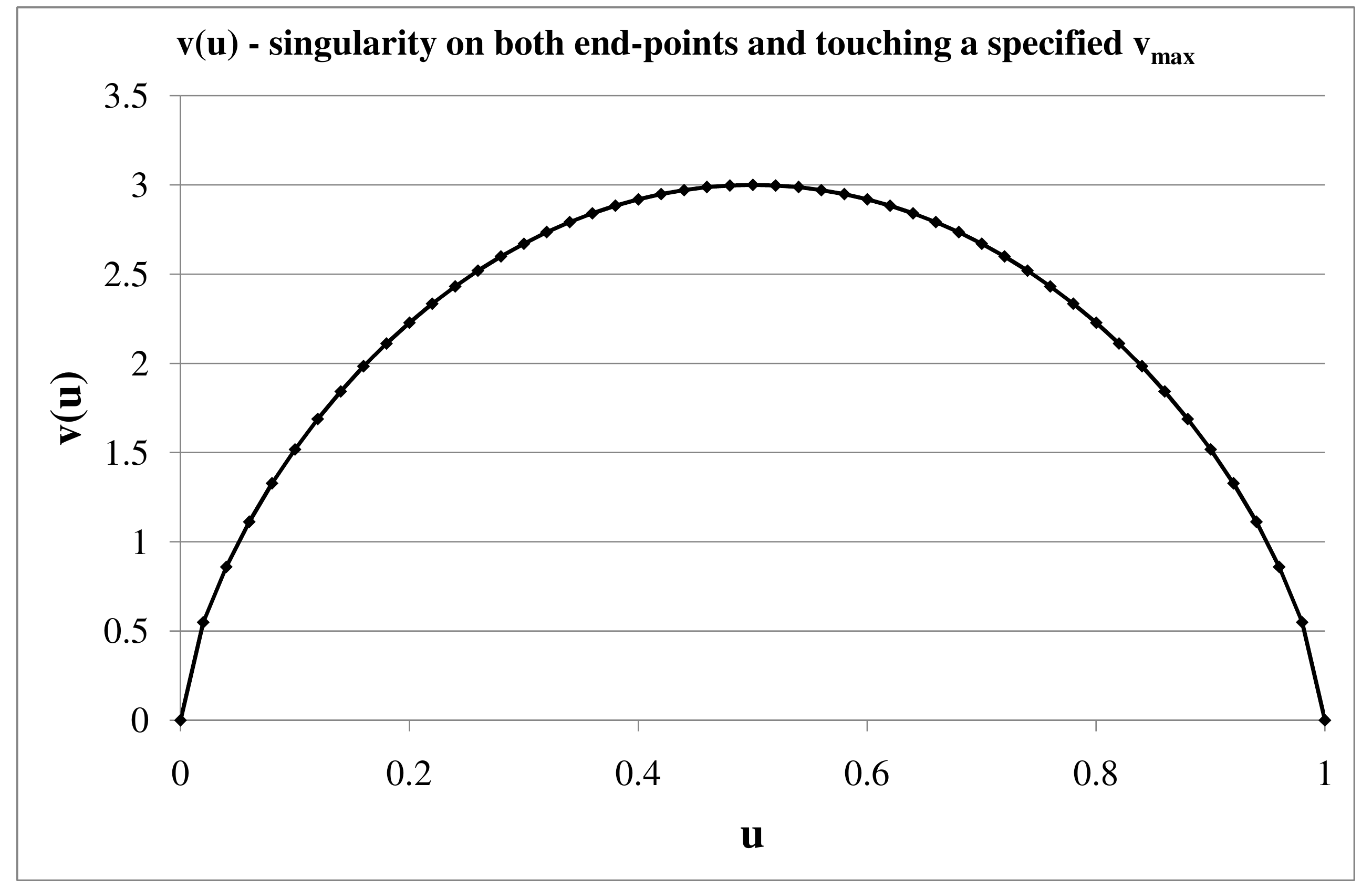}}\\
\subfigure[$v_{\max} = 3$ is active, singular start]{\label{fig:v_max_left_sing}  \includegraphics[scale=0.25]{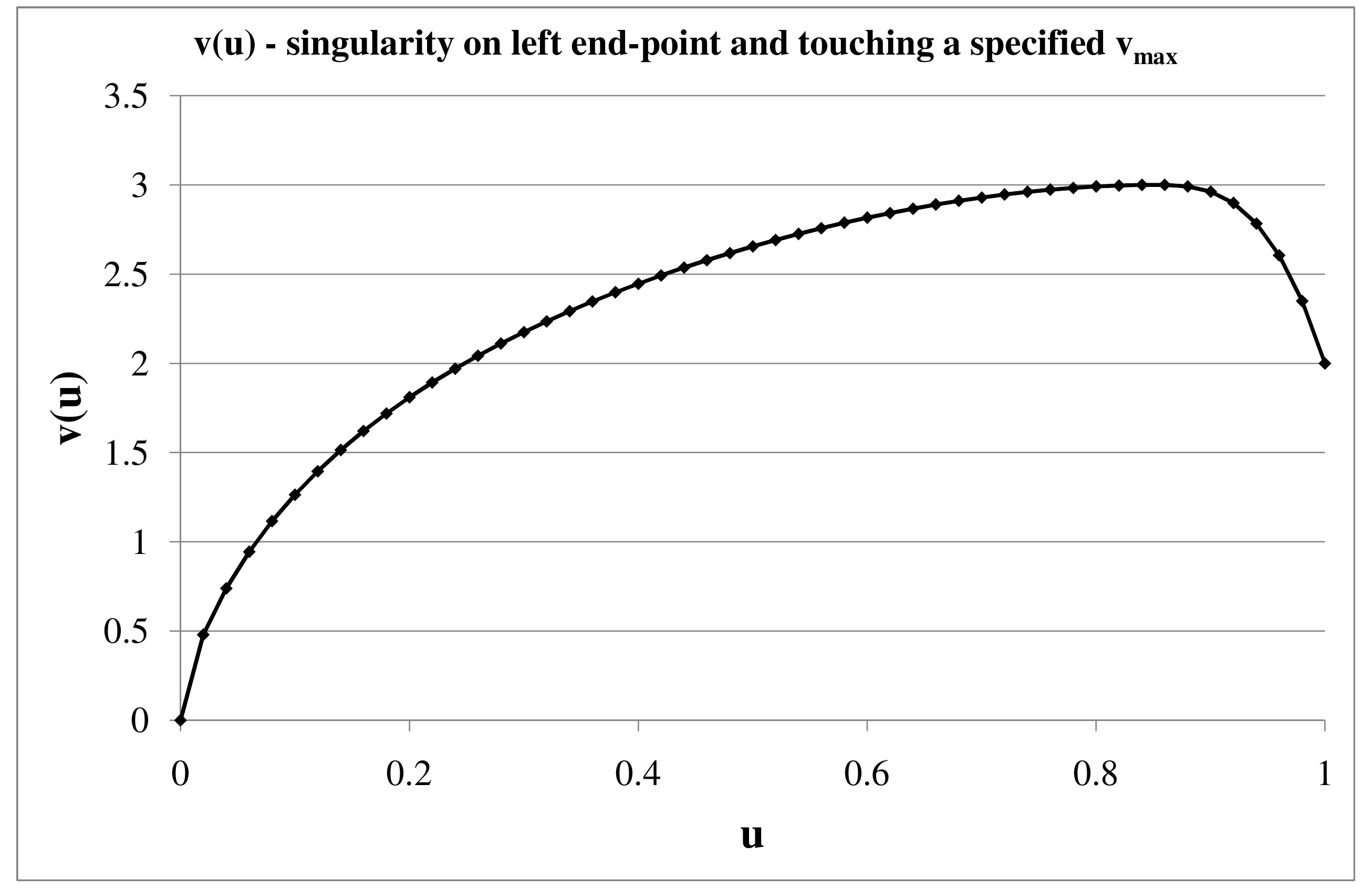}}
\subfigure[$v_{\max} = 3$ is active, singular end]{\label{fig:v_max_right_sing} \includegraphics[scale=0.25]{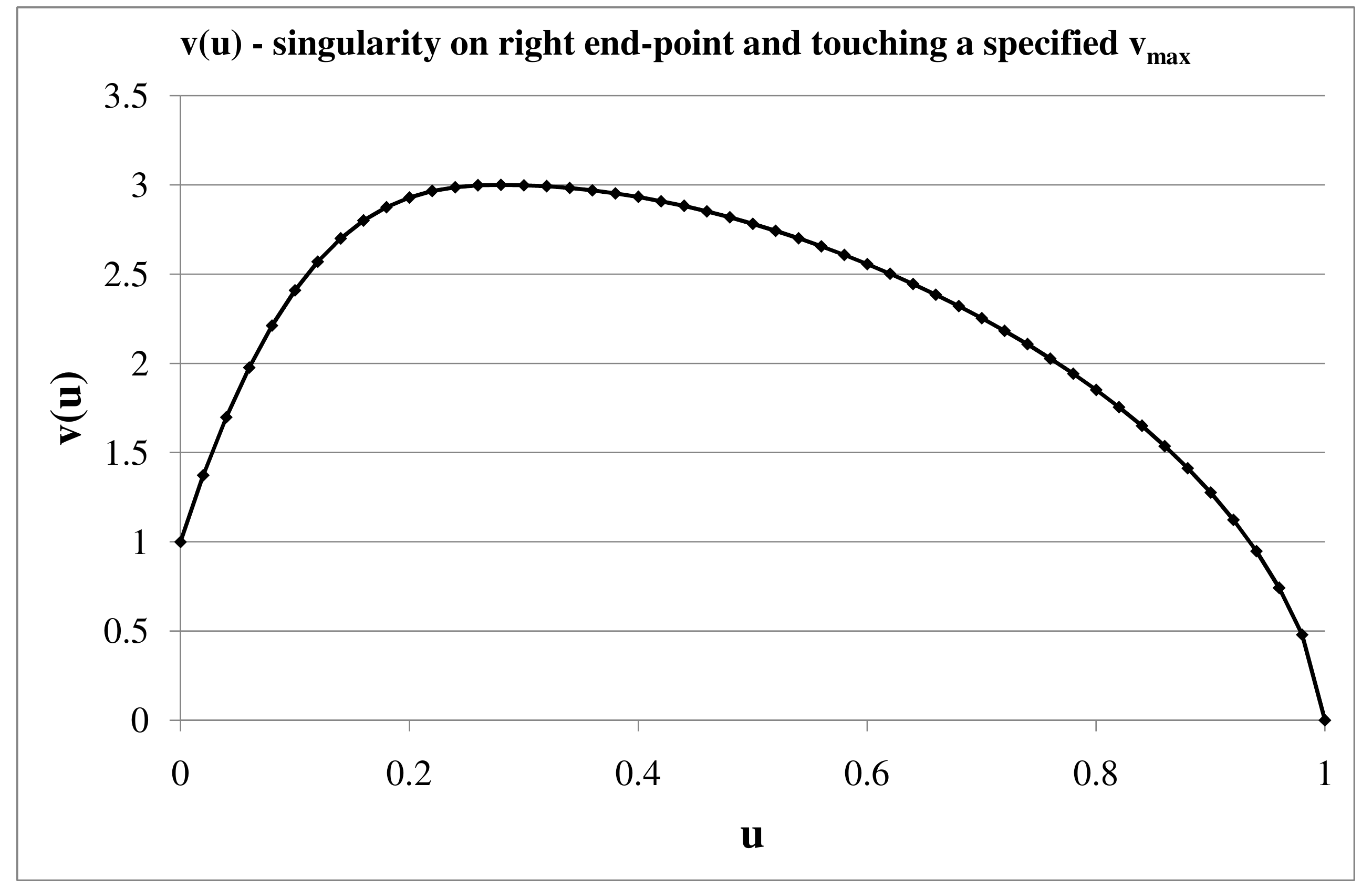}}
\end{center}
\caption{Initial guesses for speed}
\label{fig:v_init_guess}
\end{figure}

\Fig{fig:v_init_guess} shows these different cases.  All the
imposed bounds are maintained and the initial guesses of $v$
are smooth curves for all kinds of boundary conditions.  For
non-zero boundary speed, the values are 1 on the starting
point and 2 on the ending point. Maximum speed is 3.  Where
imposed, $A_{\min} = -50$ and $A_{\max} = 50$.

\clearpage

\subsection{Implementation details}
We have implemented our code in C++. We use Ipopt, a robust large-scale 
nonlinear constrained
optimization library~\citep{Wachter_2006}, also written in C++, to solve the
optimization problem to compute the initial guess of path (\Sec{sec:iguess_path_opt})
 and speed (\Sec{sec:iguess_speed}). 
We explicitly compute gradient and Hessian 
problem in our code instead of letting Ipopt compute these using finite difference.
This leads to greater robustness and faster convergence.  We set
the Ipopt parameter for relative tolerance as $10^{-8}$ and the maximum number
of iterations to $100$.

\subsection{Computational time for initial guess}

To evaluate the reliability of our method, we constructed a set of $7500$ 
problems with different boundary conditions and solved the full constrained 
optimization problem corresponding to each of the 4 initial guesses for each problem.
We present an analysis of the reliability and  run-time for the initial guesses for this 
problem set.

We generate the problem set as follows. Fix the initial position as 
$\left\{0,0 \right\}$ and orientation as 0. Choose final position
at different distances along radial lines from the origin.
Choose 10 radial lines that start from 0 degrees and go up to 
180 degrees in equal increments. The distance on the radial line
is chosen from the set $\left\{1, 2, 4, 8, 16\right\}$. The angle
of the radial line and the distance on the line determines the
final position. Choose 30 final orientations starting from 
0 up to 360 degrees (360 degrees not included) in equal increments. The 
speed, $v$, and tangential acceleration,
$a_{\tny T}$, at both ends are varied by choosing $\left\{v, a_{\tny T}
\right\}$ pairs from the set $\left\{ \left\{0, 0\right\}, \left\{1, -0.1\right\},
\left\{1, 0\right\}, \left\{1, 0.1 \right\}, \left\{3, 0 \right\} \right\}$.
Thus we have 10 radial lines, 5 distances on each radial line, 30 orientations,
5 $\left\{v, a_{\tny T} \right\}$ pairs, resulting in
$10 \times 5 \times 30 \times 5 = 7500$ cases. Each problem has $189$ degrees of freedom.

\begin{figure}[p!]
%\begin{figure}
\begin{center}
	\subfigure[99\% solved within 0.2 s.]
	{
		\includegraphics[trim = 0mm 0mm 20mm 0mm, clip, scale = 0.16] {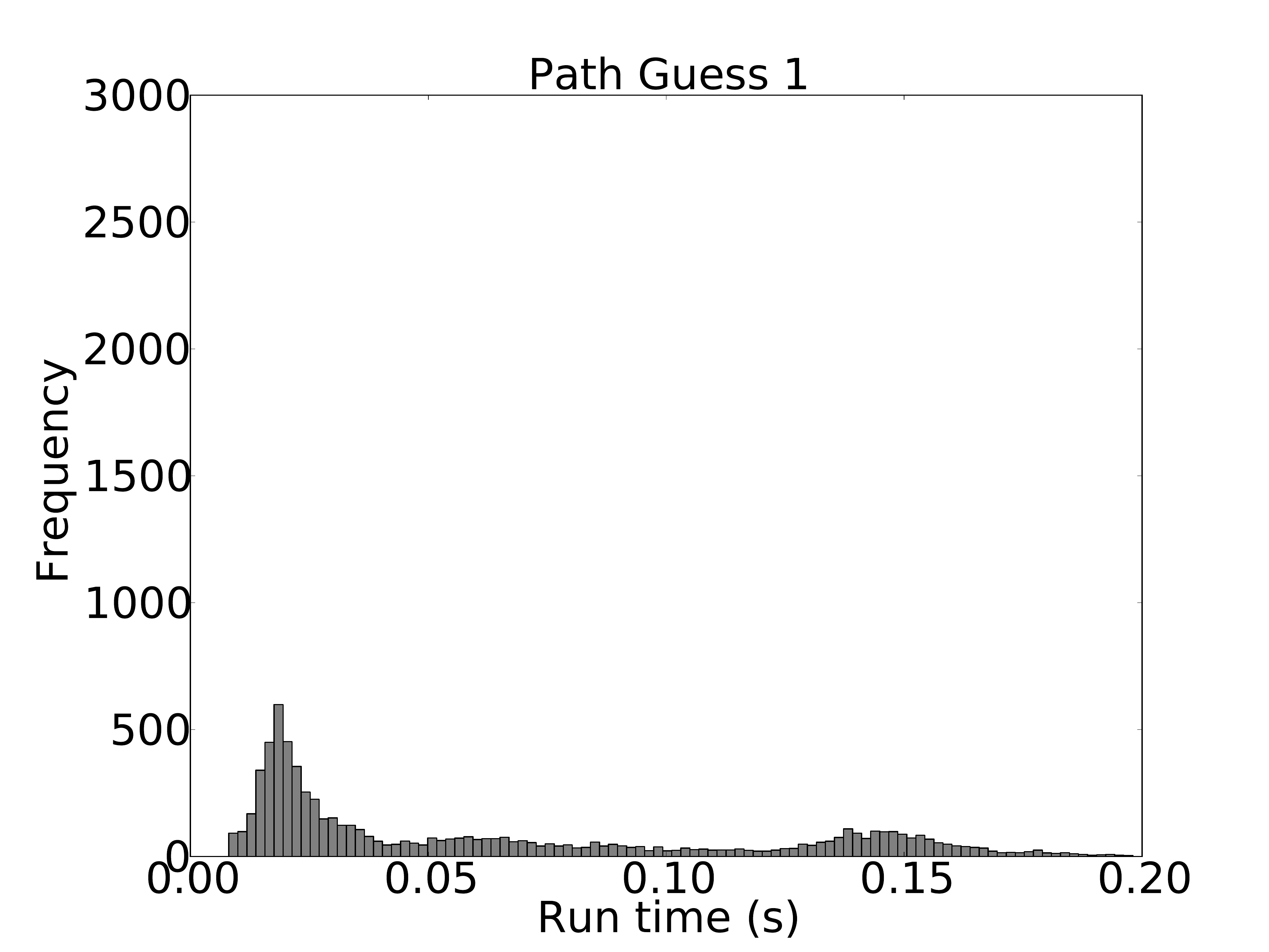}
	}
	\subfigure[99\% solved within 0.2 s.]
	{   
	    \includegraphics[trim = 0mm 0mm 20mm 0mm, clip, scale = 0.16] {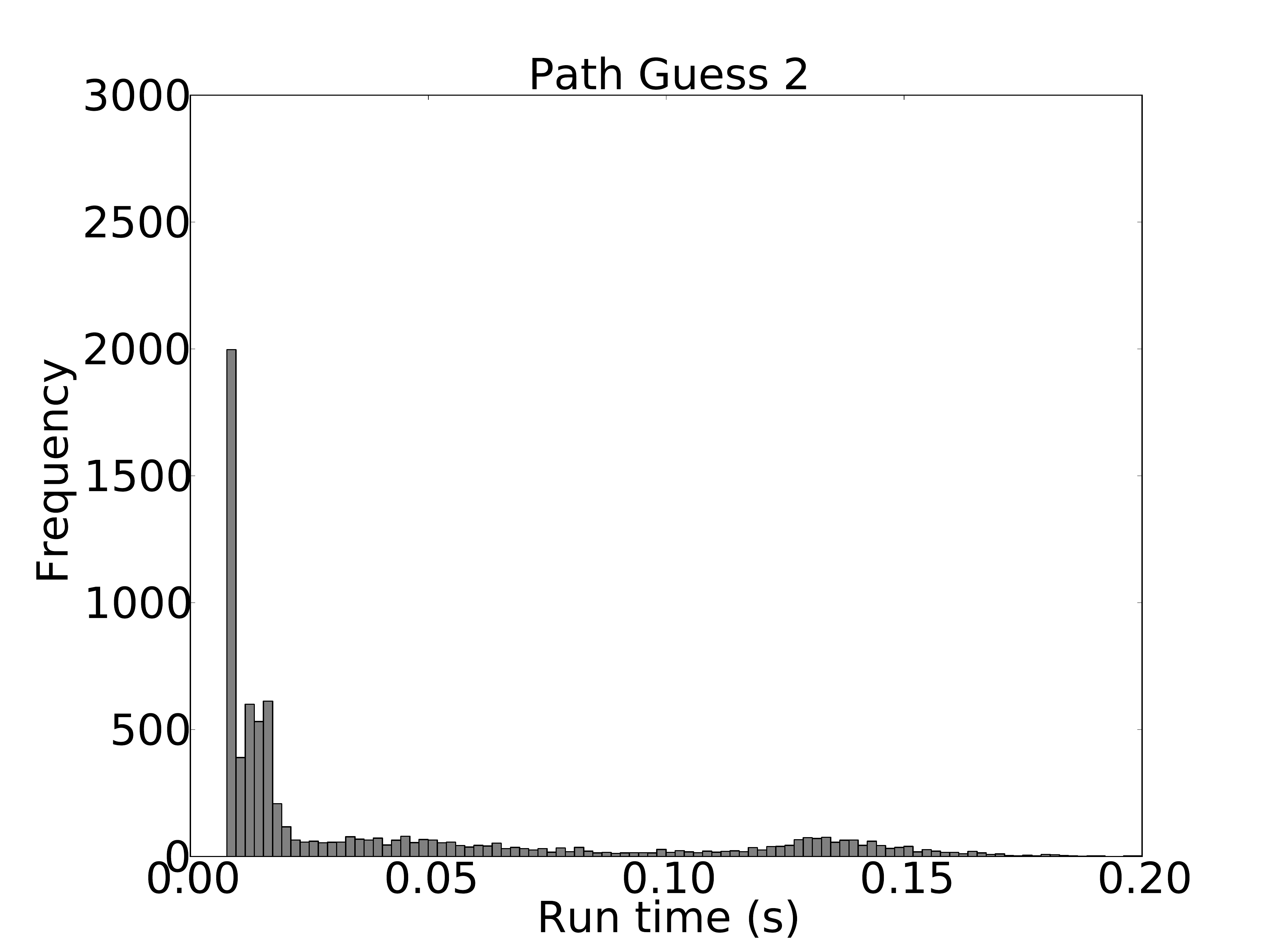}
	}\\
	\subfigure[100\% solved within 0.2 s.]
	{
	    \includegraphics[trim = 0mm 0mm 20mm 0mm, clip, scale = 0.16]
	    {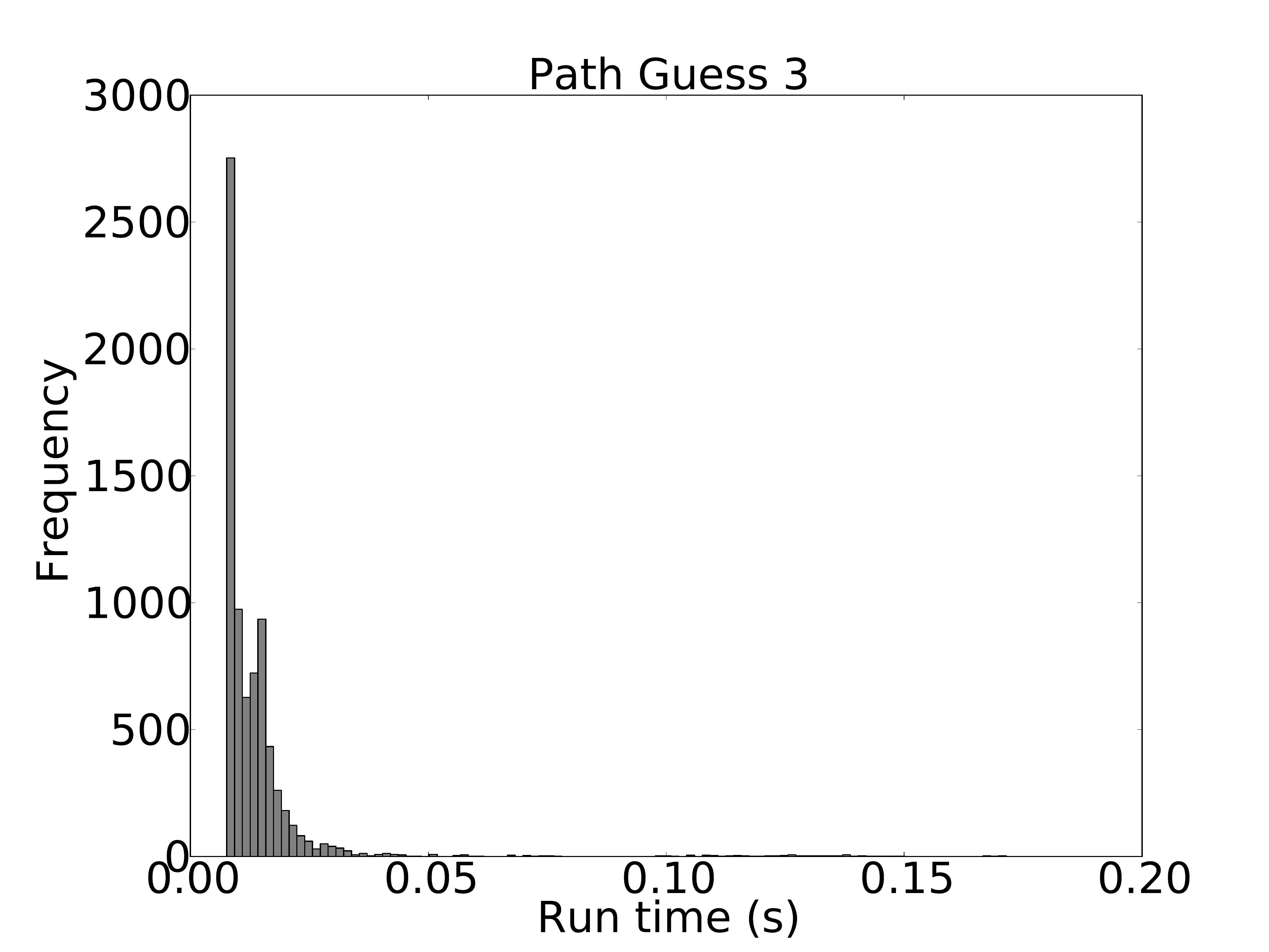}
	}
	\subfigure[100\% solved within 0.2 s.]
	{
	    \includegraphics[trim = 0mm 0mm 20mm 0mm, clip, scale = 0.16]
	    {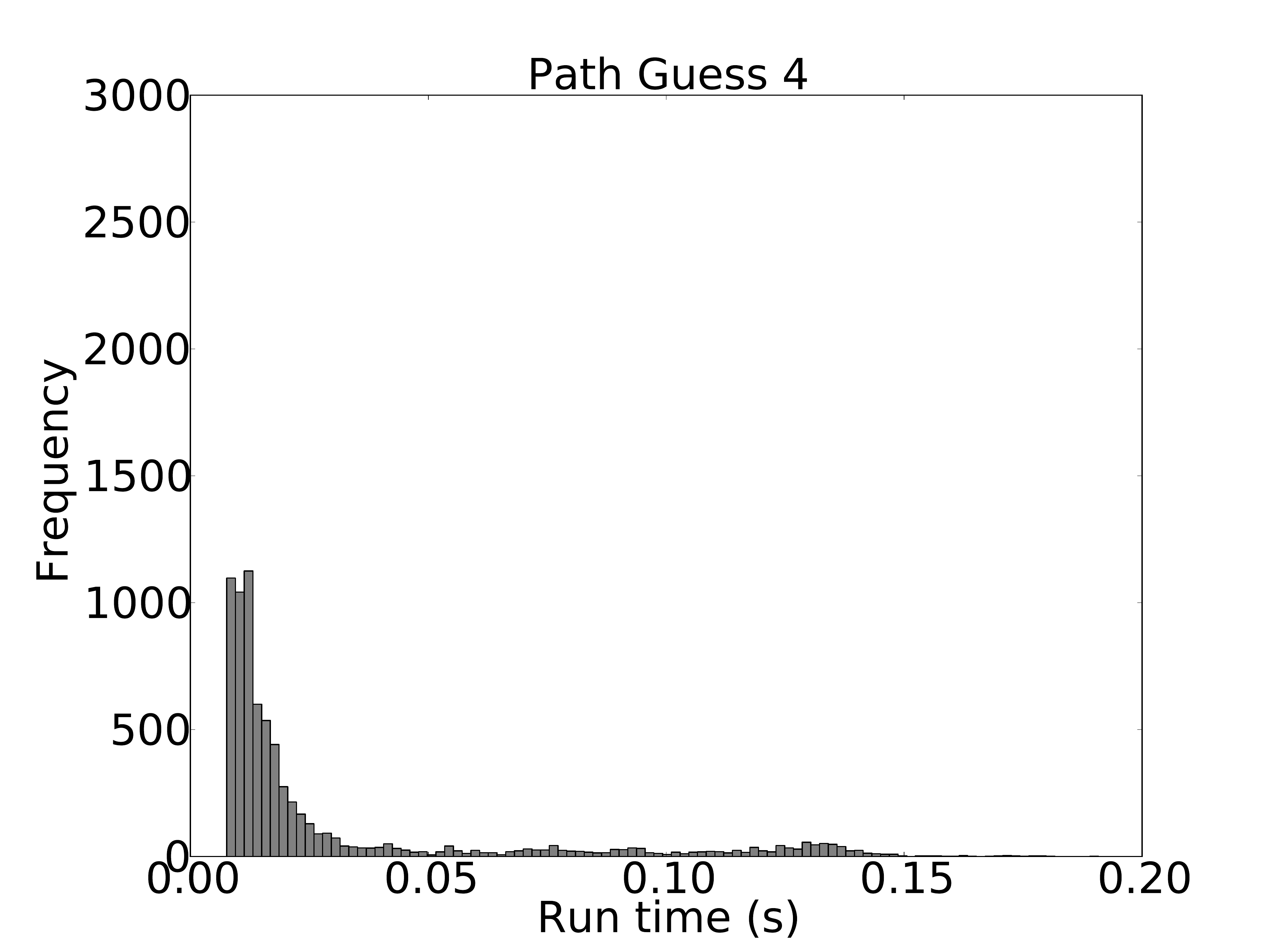}
	}
\end{center}
\caption{Histogram of time taken to compute initial guess of path. Total 7500 cases.}
\label{fig:theta_init_guess_time}
%\end{figure}

%\begin{figure}
\begin{center}
	\subfigure[][99\% solved within 0.12 s.]
	{
		\includegraphics[trim = 0mm 0mm 20mm 0mm, clip, scale = 0.16] {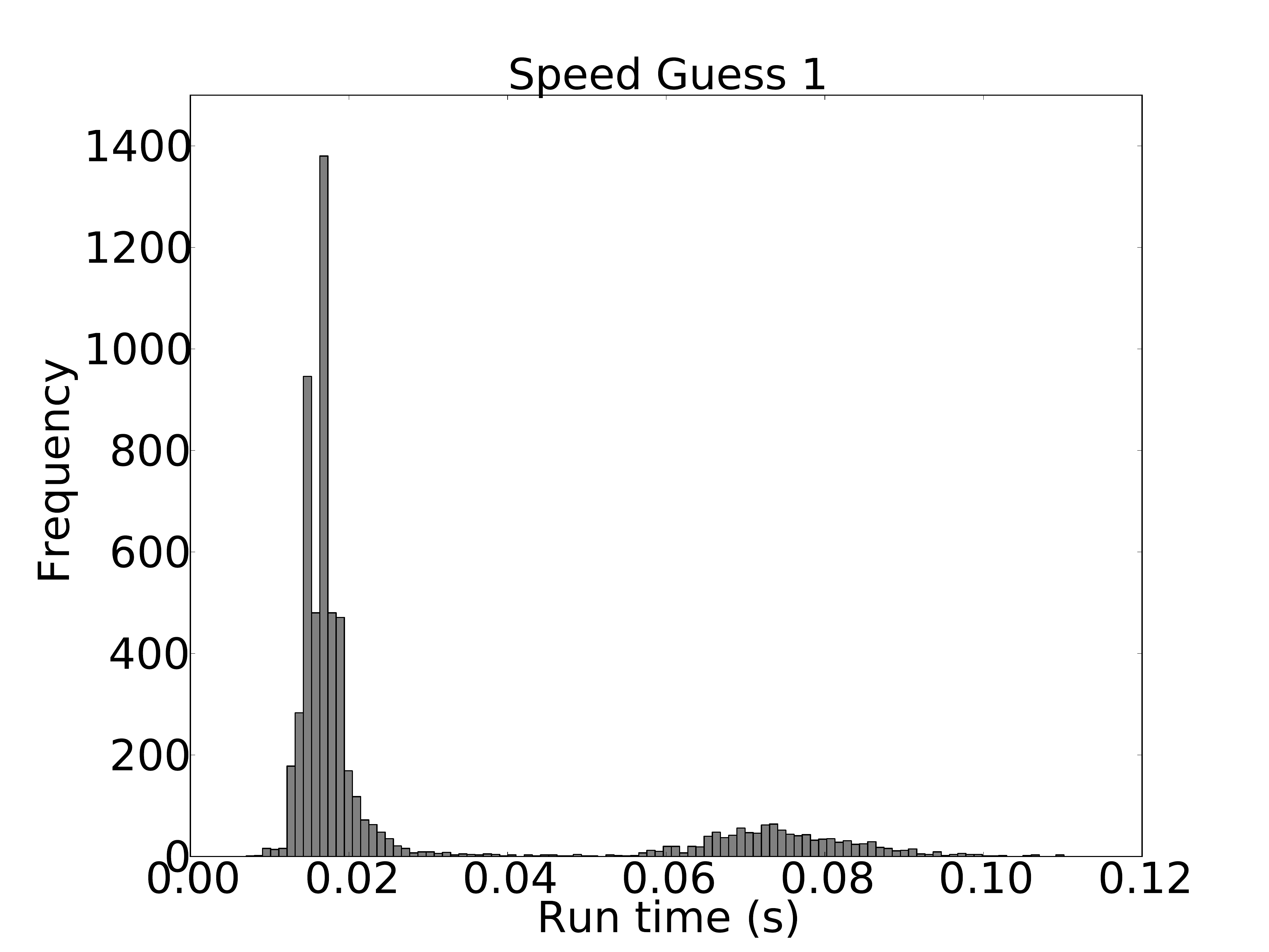}
	}
	\subfigure[][99\% solved within 0.12 s.]
	{   
	    \includegraphics[trim = 0mm 0mm 20mm 0mm, clip, scale = 0.16] {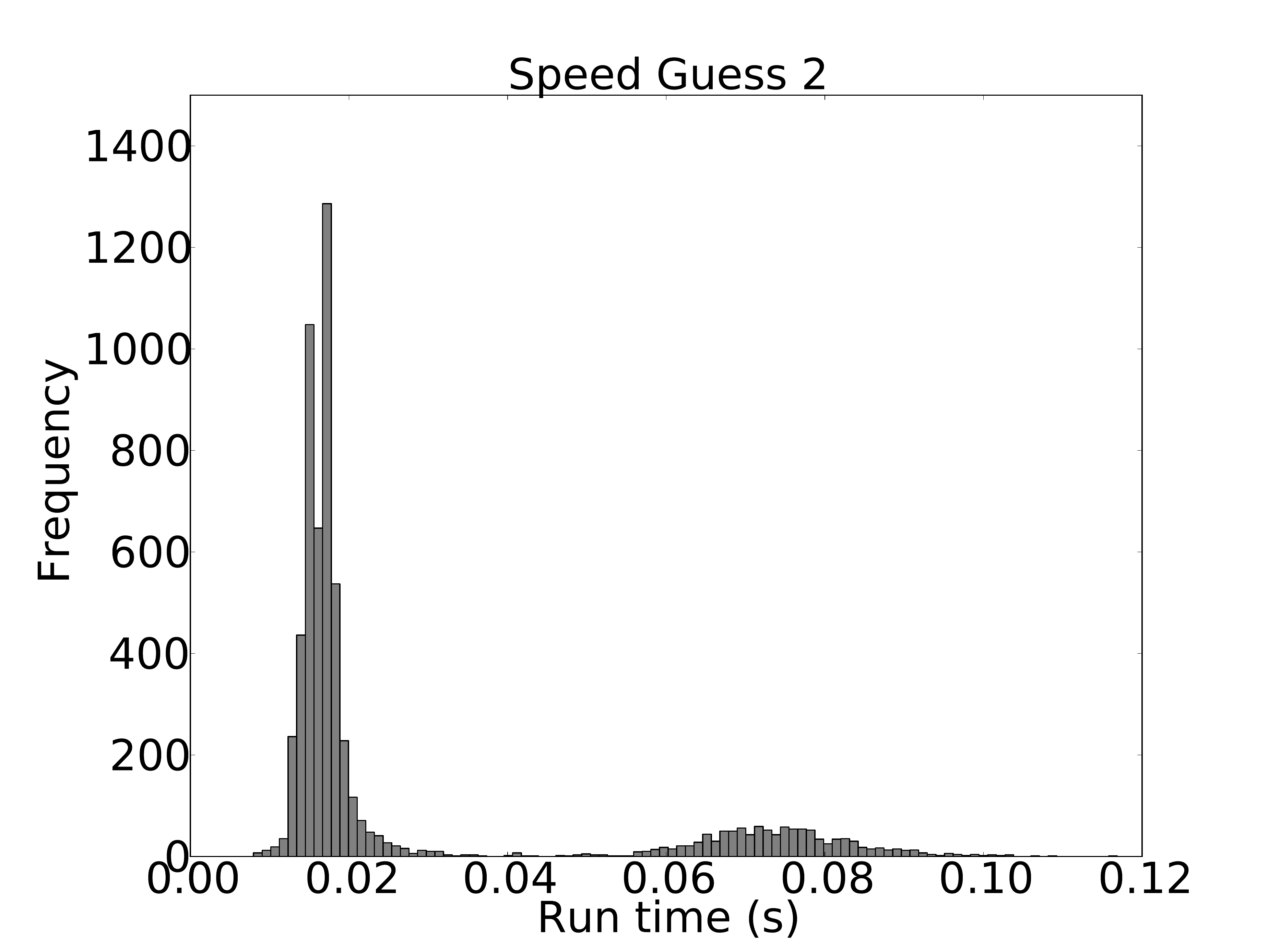}
	}\\
	\subfigure[][99\% solved within 0.12 s.]
	{
	    \includegraphics[trim = 0mm 0mm 20mm 0mm, clip, scale = 0.16]
	    {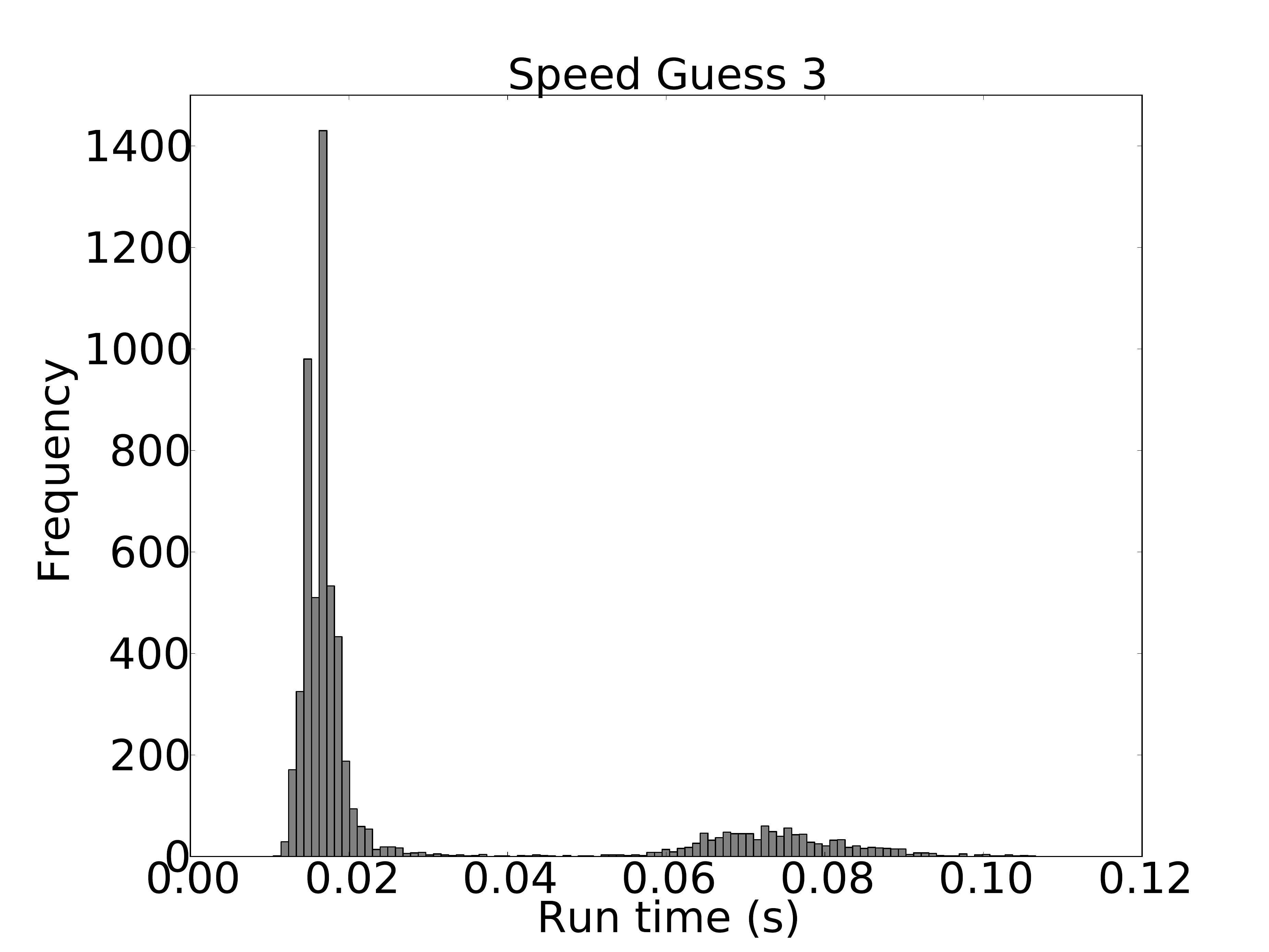}
	}
	\subfigure[][99\% solved within 0.12 s.]
	{
	    \includegraphics[trim = 0mm 0mm 20mm 0mm, clip, scale = 0.16]
	    {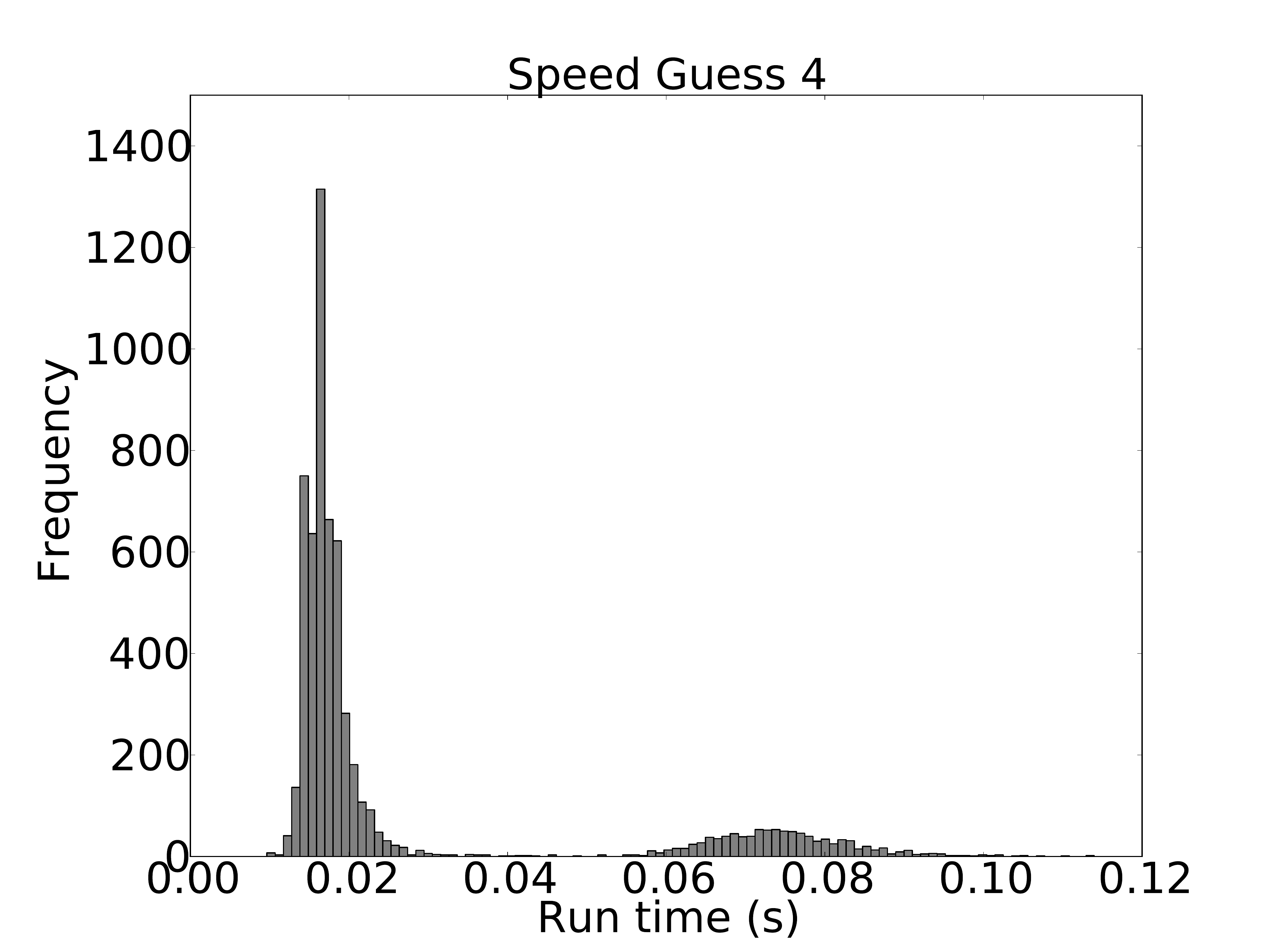}
	}
\end{center}
\caption{Histogram of time taken to compute initial guess of speed. Total 7500 cases.}
\label{fig:vel_init_guess_time}
\end{figure}

All the problems were solved on a computer with an Intel Core i7 CPU running
at 2.67 GHz, 4 GB RAM, and 4 MB L-2 cache size. Histograms of run-time for 
computing initial guess of speed and initial
guess of path are shown in Figures~\ref{fig:vel_init_guess_time} and ~\ref{fig:theta_init_guess_time} respectively. Histograms for all four initial
guesses and all four discomfort minimization problems are shown.
In all these histograms, we have removed 1\% or less of cases that lie 
outside the range of the axis shown for better visualization. 
All histograms show both successful and unsuccessful
cases. 

For all 7500 problems, at least one initial guess was successfully 
computed. From \Fig{fig:theta_init_guess_time} we see that than 99\% or
more of initial guesses of path are computed in less than 0.2 s. 
From \Fig{fig:vel_init_guess_time} we see that 99\% or more
of initial guesses of speed are computed in less than 0.12 s.

These initial guesses were used to solve the full nonlinear optimization 
problem. Results for the full problem are presented in Part I.
Here is a summary. Out of a set of $7500$ examples with
varying boundary conditions, and all dynamic constraints
imposed, 3.6 solution paths, on average, were found per example. 
At least one solution was found for all of the problems and
four solutions were found for roughly 60\% of the problems.
The time taken to compute the solution to the discomfort
minimization problem was less than 10 seconds for all
the cases, 99\% of all problems were solved in less
than 4 seconds, and roughly 90\% were solved in less
than 100 iterations.

\clearpage

\begin{figure}[h!]
\begin{center}
	\subfigure[]
	{
		\includegraphics[trim = 0mm 0mm 35mm 0mm, clip, scale = 0.16] {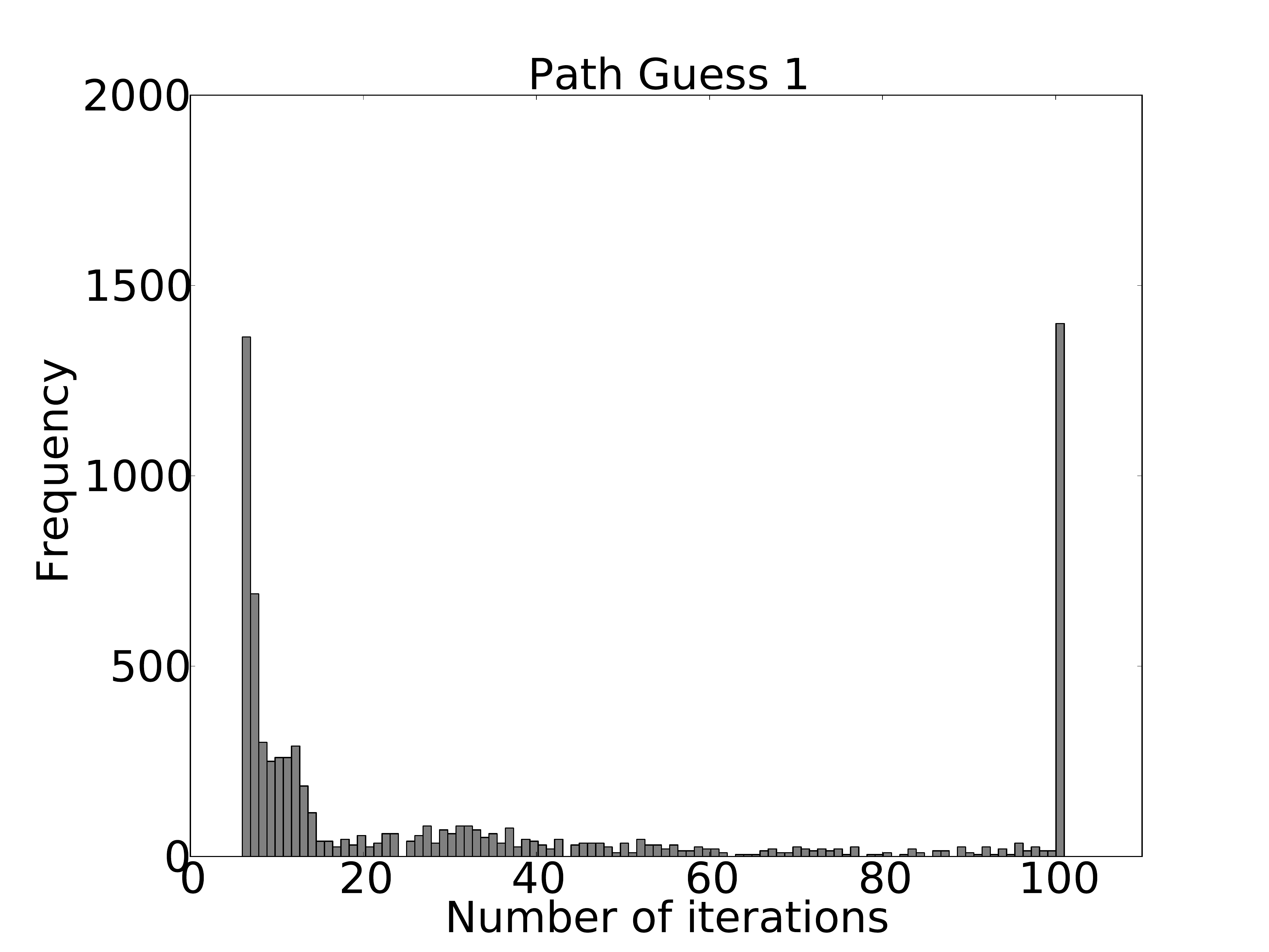}
	}
	\subfigure[]
	{   
	    \includegraphics[trim = 0mm 0mm 35mm 0mm, clip, scale = 0.16] {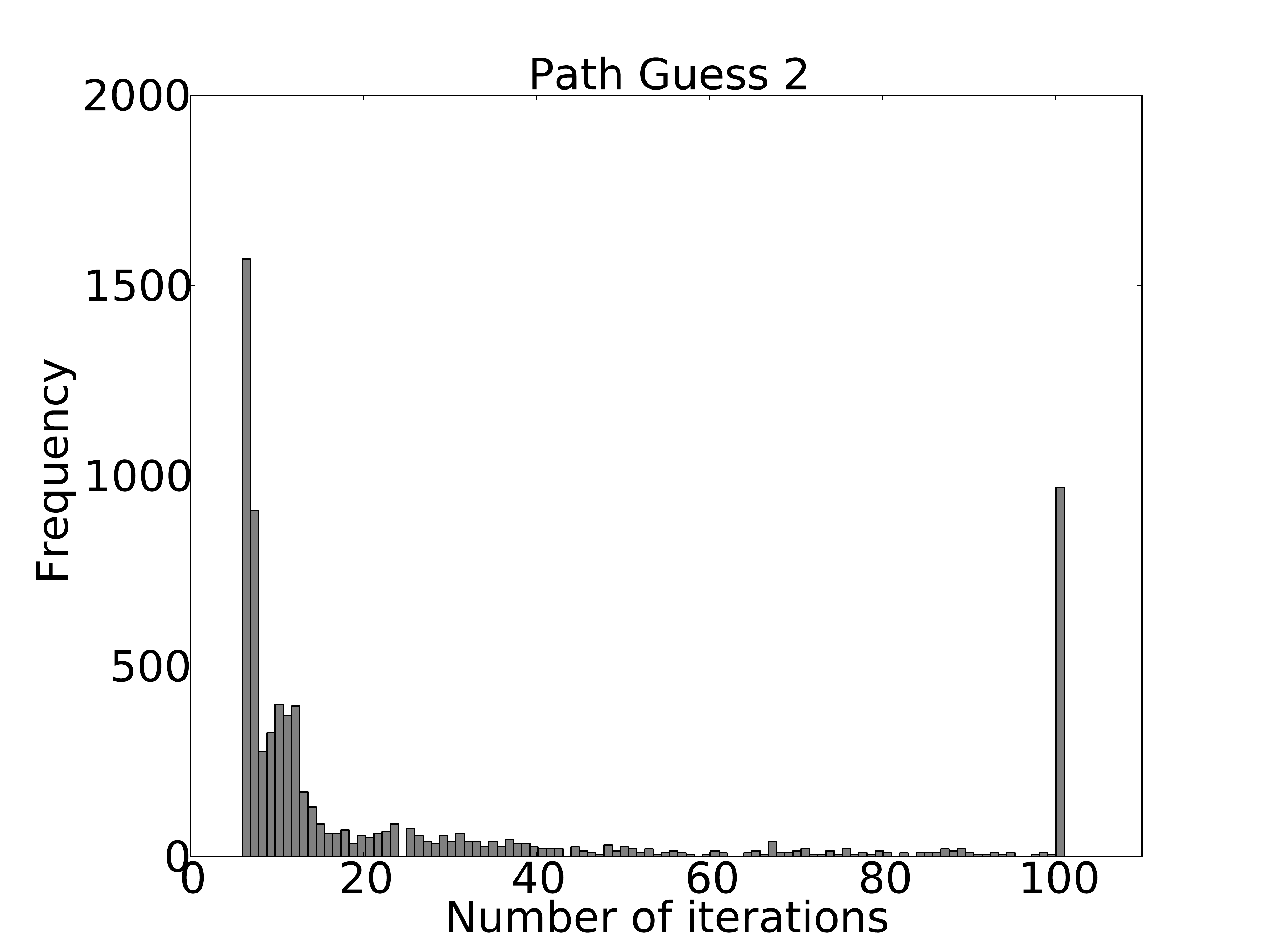}
	}\\
	\subfigure[]
	{
	    \includegraphics[trim = 0mm 0mm 35mm 0mm, clip, scale = 0.16]
	    {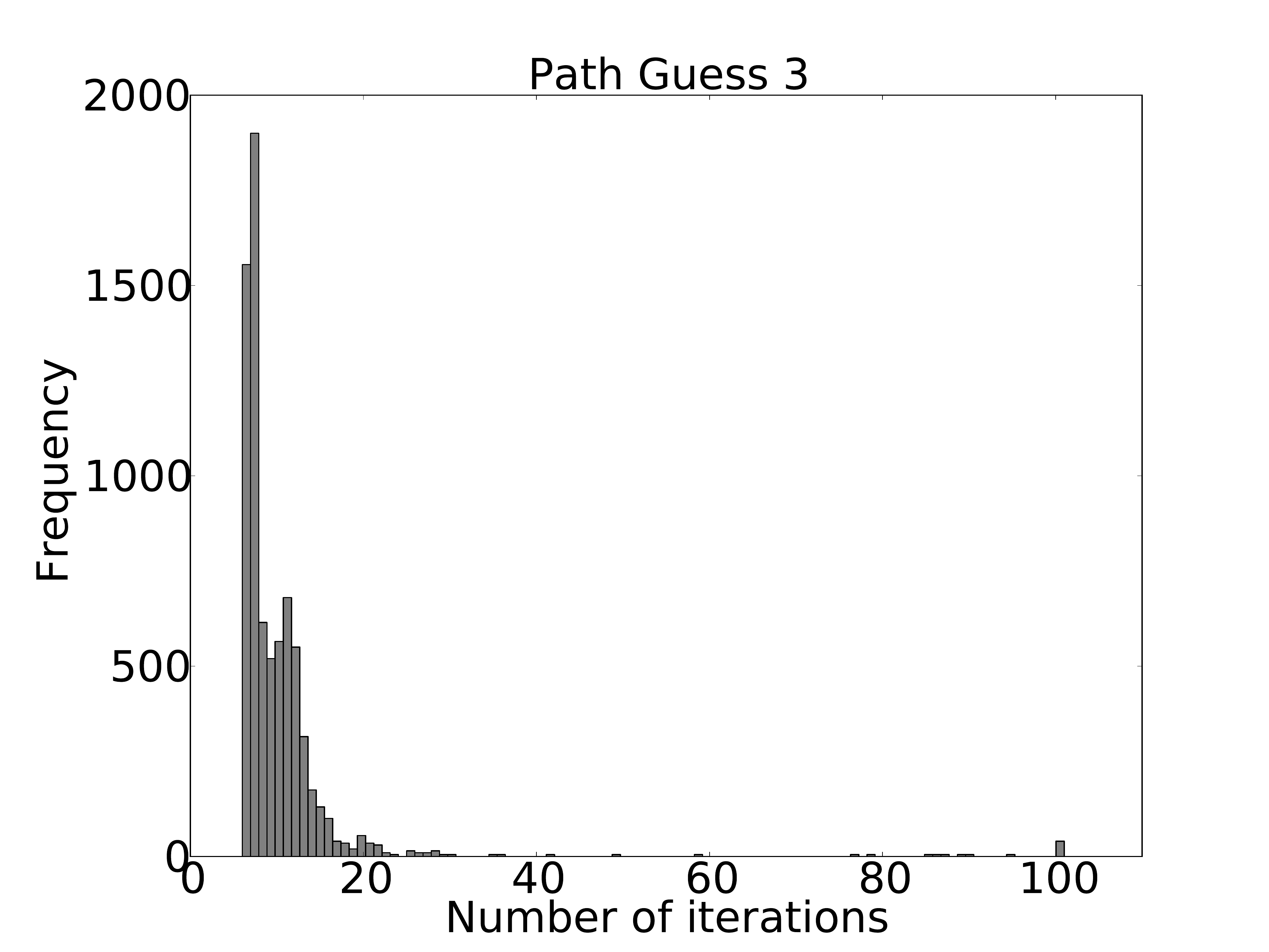}
	}
	\subfigure[]
	{
	    \includegraphics[trim = 0mm 0mm 35mm 0mm, clip, scale = 0.16]
	    {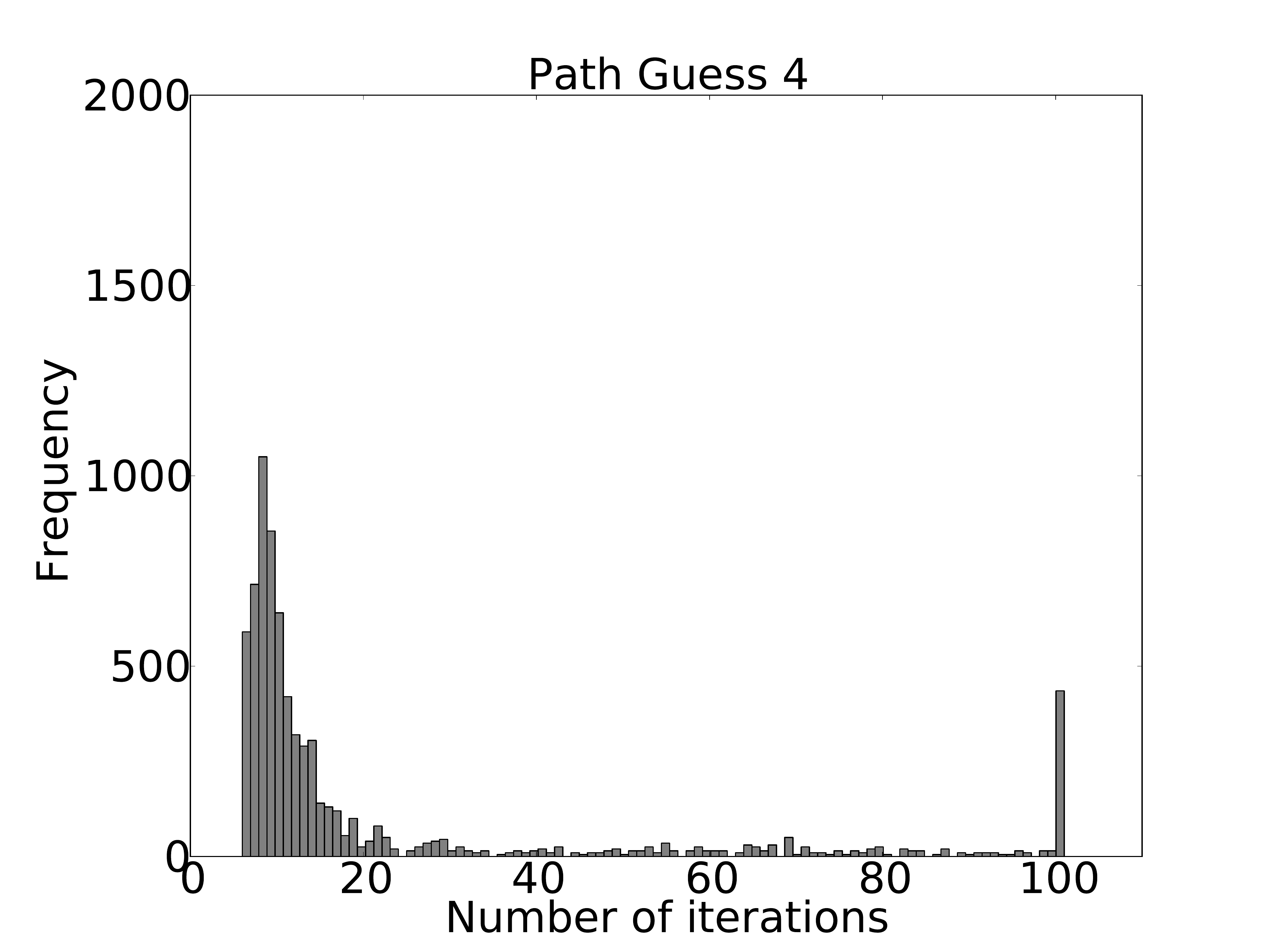}
	}
\end{center}
\caption{Histogram of number of iterations to compute initial guess of path.}
\label{fig:theta_init_guess_iters}
\end{figure}

\clearpage

%\section{Concluding remarks}
%In this series of papers, we developed a nonlinear constrained optimization 
%based motion planning framework for planning comfortable trajectories for a nonholonomic
%robot moving on plane. In the first part of this series, we presented
%the mathematical foundation of the framework and formulated the
%full infinite-dimensional nonlinear optimization problem with all
%the boundary conditions and constraints. In this paper, we presented
%a numerical solution method for the problem. We discretized the problem 
%using the finite element method and described the full finite
%dimensional optimization problem. We also described a method to compute 
%good quality initial guesses for the optimization problem. Our results 
%show that our approach is
%capable of reliably planning trajectories for a wide range of boundary 
%conditions such that the trajectories satisfy all the properties
%necessary for comfort. We believe 
%that our work is a significant step toward planning comfortable motion for 
%human users of autonomous mobile robots.

\section{Acknowledgements}
This work has taken place in the Intelligent Robotics Lab
at the Artificial Intelligence Laboratory, The University of
Texas at Austin. Research of the Intelligent Robotics lab
was supported in part by grants from the National Science
Foundation (IIS-0413257, IIS-0713150, and IIS-0750011), 
the National Institutes of Health (EY016089), and from the 
Texas Advanced Research Program (3658-0170-2007).

\bibliographystyle{apalike}
\bibliography{ijrr_2012_part2}

\end{document}